\documentclass[10pt,twocolumn,letterpaper]{article}

\usepackage{cvpr}              

\usepackage{xspace}
\newcommand{\method}{Match-and-Fuse\xspace}
\newcommand{\mff}{Multiview Feature Fusion\xspace}
\newcommand{\mguide}{Feature Guidance\xspace}
\newcommand{\graph}{Pairwise Consistency Graph\xspace}
\newcommand{\afterfigure}{\vspace{-5mm}}  

\usepackage[ruled]{algorithm2e} 
\SetEndCharOfAlgoLine{}
\SetCommentSty{mycommfont}
\SetKwComment{tcc}{\hfill$\triangleright$~}{}  
\SetKwComment{tcp}{\hfill$\triangleright$~}{}  

\SetKwInput{KwInput}{Inputs}
\SetKwInput{KwHyper}{Hyperparameters}
\SetKwInput{KwAlgo}{Algorithm}
\SetKwInput{KwOutput}{Output}

\SetAlFnt{\small}
\SetAlCapFnt{\small}
\SetAlCapNameFnt{\small}
\SetAlCapHSkip{0pt}
\usepackage{float}     

\newcommand{\myparagraph}[1]{\vspace{0.15cm}\noindent{\it #1}\hspace{0.05cm}} 

\definecolor{cvprblue}{rgb}{0.21,0.49,0.74}
\usepackage[pagebackref,breaklinks,colorlinks,allcolors=cvprblue]{hyperref}

\def\paperID{3204}
\def\confName{CVPR}
\def\confYear{2026}

\title{\method: Consistent Generation from Unstructured Image Sets}

\author{Kate Feingold \hspace{8mm} Omri Kaduri \hspace{8mm} Tali Dekel \\\\
Weizmann Institute of Science}

\begin{document}
\twocolumn[{
  \maketitle
  \centering
  \vspace{-0.5cm}
  \includegraphics[width=\textwidth]{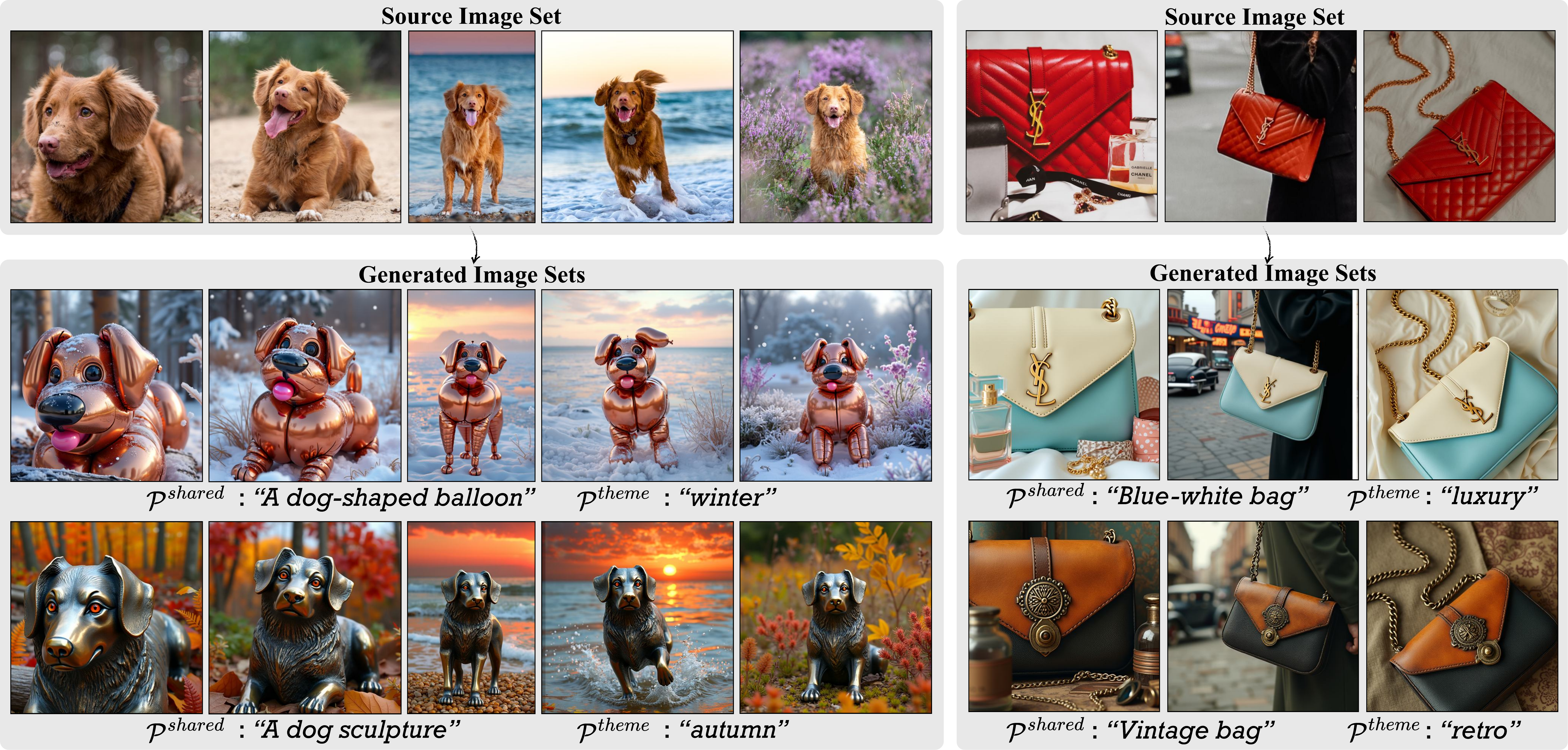}
  \vspace{-2em}
  \captionof{figure}{Given a source set of images (top), depicting shared objects in varied settings (\emph{e.g.}, pose, environment, viewpoint), our method, \textbf{\method}, jointly generates an output set in which the consistency among the shared content is preserved (bottom). The output adheres to the user-provided prompts that describe the target shared content ($\mathcal{P}^\textit{shared}$), and the scene's style/theme ($\mathcal{P}^\textit{theme}$). }
  \label{fig:teaser}
  \vspace{1em}
}]

\begin{abstract}
We present \method{} -- a zero-shot, training-free method for consistent controlled generation of unstructured image sets -- collections that share a common visual element, yet differ in viewpoint, time of capture, and surrounding content. Unlike existing methods that operate on individual images or densely sampled videos, our framework performs set-to-set generation: given a source set and user prompts, it produces a new set that preserves cross-image consistency of shared content. Our key idea is to model the task as a graph, where each node corresponds to an image and each edge triggers a joint generation of image pairs. 
This formulation consolidates all pairwise generations into a unified framework, enforcing local consistency while ensuring global coherence across the entire set. This is achieved by fusing internal features across image pairs, guided by dense input correspondences, without requiring masks or manual supervision, and by leveraging an emergent prior in text-to-image models that encourages coherent generation when multiple views share a single canvas.
\method{} achieves state-of-the-art consistency and visual quality, and unlocks new capabilities for content creation from image collections. 
Code and data: \href{https://match-and-fuse.github.io/}{project page}.
\end{abstract}

\vspace{-1cm}
\section{Introduction}

Much of our visual experience—how we capture, organize, and interpret the world—is structured not around single images but around sets of them. Photo albums, real-estate listings, product catalogs, and historical archives all offer multiple perspectives on shared content captured at different times or viewpoints. Such sets are richer than individual images yet more flexible than continuous video. Despite major progress in single-image and video generation, Generative AI remains largely underexplored for this fundamental unit of visual communication.

Given a source image set and a user prompt describing the desired shared content, our method produces an output set that adheres to the prompt while preserving the source layout and the cross-image consistency of shared elements, as illustrated in \cref{fig:teaser}. These shared elements—semantic regions appearing across multiple images—are kept coherent in identity, appearance, and geometry. Importantly, non-shared regions (e.g., backgrounds) are not forced to align and may vary according to a separate thematic prompt. This capability enables creative workflows of transforming fixed multi-view layouts into coherent edits for product ads, character concept art, film set design, and more.

Achieving consistency across an image set is nontrivial. Unlike videos, which benefit from dense temporal sampling and continuity, image sets challenge this assumption. Without temporal cues such as continuous motion, enforcing visual consistency becomes difficult. The problem is further compounded when the content is deformable -- e.g., varying human poses, expressions, or environments -- where no stable 3D structure can be inferred. Consequently, existing generative methods lack mechanisms to model shared content without strong spatial or temporal cues.

Our method builds on a pre-trained text-to-image diffusion model in a zero-shot, training-free manner. Our key idea is to model an image set as a graph whose nodes are images and whose edges represent joint pairwise generation. This formulation consolidates all pairwise generations into a joint framework, achieving coherence both locally between image pairs and globally across the entire set.
Crucially, this formulation allows us to: (i) harness an emergent prior in text-to-image diffusion models, which exhibit coherent behavior when multiple images are composed within a shared generation canvas; (ii) enhance consistency by incorporating dense pixel correspondences computed from the source images, guiding alignment without requiring masks or manual supervision; (iii) flexibly handle sets of varying sizes without sacrificing resolution. We demonstrate the effectiveness of our framework in producing high-quality results from diverse collections of 2–15 images, ranging from real photographs to hand-drawn sketches. 

To summarize, we make the following contributions: 
\begin{itemize}
\item The first method for unstructured \emph{set-to-set generation}, moving beyond image pairs to entire collections.
\item  A flexible, automated, training-free mask-free approach, requiring only simple text prompts as input.
\item A new metric for evaluating fine-grained cross-image consistency that aligns strongly with human judgments.
\end{itemize}

\begin{figure*}[t!]
  \centering
  \includegraphics[width=1.0 \linewidth,keepaspectratio]{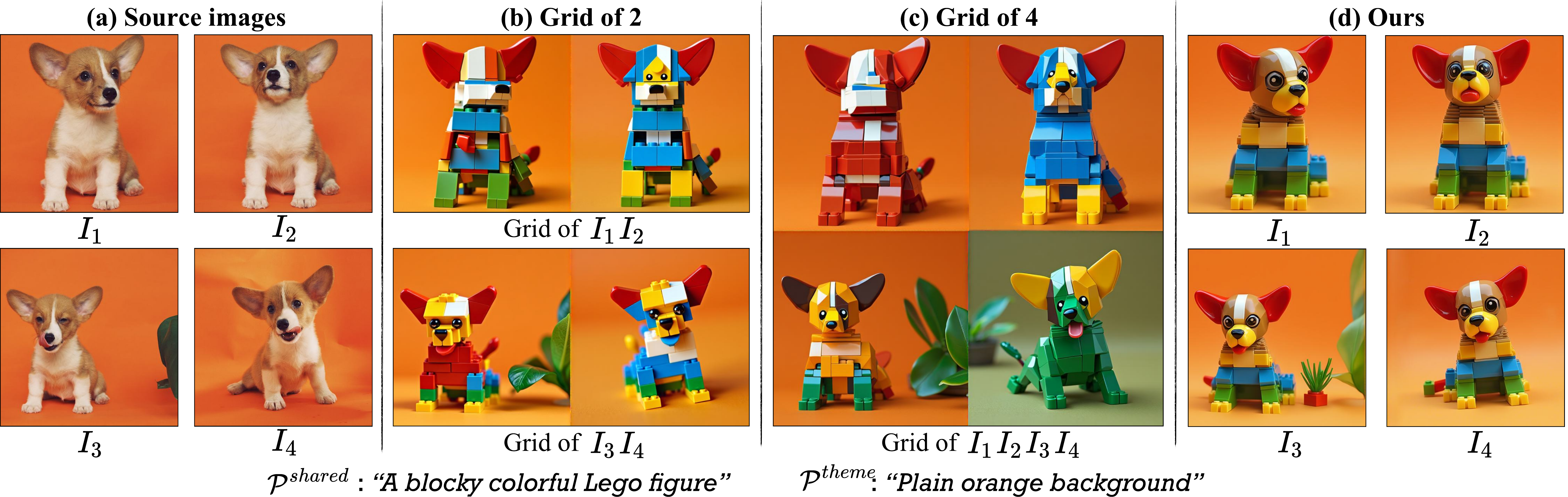}%
  \vspace{-7pt}
  \caption{\textbf{Image grid generation vs. our method.}
    (a) Source image set. (b) Joint generation of two-image grid results in partial consistency, where several regions often remain inconsistent in appearance or semantic meaning (e.g., a dog's face). 
    (c) Extending this to more images further reduces consistency. (d) Our method leverages this prior yet overcomes its limitations of consistency and scale. 
  } \afterfigure
  \label{fig:grid_prior}
\end{figure*}

\vspace{-5pt}
\section{Related work}
\vspace{-5pt}
\label{sec:related_work}

\myparagraph{{\bf Text-to-image controlled generation.}}
T2I models \cite{rombach2022highresolutionimagesynthesislatent,saharia2022photorealistictexttoimagediffusionmodels,shi2020improvingimagecaptioningbetter,fluxcontrolnet2024,esser2024scalingrectifiedflowtransformers} have made a remarkable progress.  Numerous methods extended T2I models to go beyond input text and condition the generation on various signals, such as style (e.g., \cite{ye2023ip-adapter,mou2023t2i}) or various spatial controls: pose, depth, or edge maps  \cite{zhang2023addingconditionalcontroltexttoimage}.
A growing line of work tackles image editing rather than conditioning, including trainable editing models \cite{brooks2022instructpix2pix,labs2025flux1kontextflowmatching,google2025nanobanana} that learn to modify an image according to text or paired examples, as well as training-free editing approaches \cite{Tumanyan_2023_CVPR,hertz2022prompt,zhu2025kv} that manipulate diffusion trajectories at inference time.
However, all these methods operate on individual images and lack the set-level consistency mechanisms required for our task.

\myparagraph{{\bf Beyond single-image generation.}}

\myparagraph{Storyboard generation.}
This line of work aims to produce image sequences depicting consistent recurring characters across frames \cite{tewel2024training,avrahami2024chosen,ye2023ip-adapter,garibi2025tokenverseversatilemulticonceptpersonalization}, or emphasizing narrative coherence across prompts \cite{zhou2024storydiffusion,Liu_2024_CVPR,he2024dreamstory,lhhuang2024iclora}.
Zero-shot methods manipulate attention features through injected correspondences or masks \cite{tewel2024training,zhou2024storydiffusion}, whereas training-based approaches fine-tune diffusion models for improved consistency \cite{he2024dreamstory,Liu_2024_CVPR,lhhuang2024iclora}.
Aside from the limited pose control offered by \cite{tewel2024training}, these models rely solely on text and lack spatial control mechanisms necessary for set-to-set generation.

\myparagraph{Consistent image-set generation.}
Related to our work, Edicho \cite{bai2024edicho} addresses pairwise consistent editing, first modifying one image and then transferring the edit to its partner using 2D correspondences to warp intermediate features. Their method is similar to ours in leveraging explicit matches and operating on unstructured image sets, but it remains strictly pairwise: edits are propagated only from a single reference image, causing coherence to degrade for views farther from that anchor.
In contrast, \cite{lhhuang2024iclora} enforces consistency by generating all images simultaneously on a single canvas, where a LoRA module \cite{hu2022lora} is fine-tuned on the multi-image prompt. This design ties the method to a fixed grid, limiting its scalability to larger image sets.

\textit{3D and video-based approaches.}
Distinct from these methods, 3D editing techniques \cite{instructnerf2023,vicanerf2023,wang2024view,gaussctrl2024} enable manipulation of static multi-view sets through rendered views but operate on 3D representations rather than directly editing 2D images. This imposes restrictive assumptions that limit their applicability to image sets with unknown camera parameters, articulated objects, and varying backgrounds.
Similarly, video editing and generation methods (\emph{e.g.}, \cite{tokenflow2023,yuan2024instructvideo,wang2024cove}) are unsuitable for our task, as they assume temporal continuity and rely on motion coherence that does not hold across unordered image sets.


\vspace{-5pt}
\section{Method}
\label{sec:method}

The input to our method is an \emph{unstructured} set of $N$ images along with user-provided prompts: $\mathcal{P}^\text{shared}$ and $\mathcal{P}^\text{theme}$, describing the target shared content and general style or theme, respectively. Our method outputs $N$ images that preserve the source semantic layout while ensuring visual consistency across shared elements.

We build on a pre-trained, frozen, depth-conditioned T2I model. Although designed for single-image generation, these models have been shown to produce image grids when prompted with joint layouts (e.g., ``Side-by-side views of...''), establishing cross-image relationships as demonstrated in recent work \cite{lhhuang2024iclora,shin2024diptychprompting}.  However, this emerged capability, which we refer to as the \emph{grid prior}, exhibits several key limitations: (i) it provides only partial consistency in appearance, shape, and semantics (\cref{fig:grid_prior}b); (ii) the consistency deteriorates rapidly as more images are composed (\cref{fig:grid_prior}b); and (iii) generating a single canvas is bounded by the model’s native resolution, limiting scalability.

Our method leverages the grid prior while overcoming its core limitations (\cref{fig:grid_prior}d).
Specifically, we model the image set as a \textbf{\graph}, comprising all possible two-image grid generations (\cref{sec:graph}). This allows us to exploit the strong inductive bias of the grid prior, while eliminating its scale limitation. To enhance visual consistency both within each image grid and across grids, we perform joint feature manipulation across all pairwise generations.  To this end, we utilize dense 2D correspondences from the source set to automatically identify shared regions -- without requiring object masks -- and enforce fine-grained alignment. We find that feature-space similarity along these matches correlates strongly with visual coherence, motivating the use of \textbf{\mff} (\cref{sec:mff}). We further refine details via  \textbf{\mguide}, using a feature-matching objective (\cref{sec:guide}). Our full pipeline is illustrated in \cref{fig:method}.

\subsection{Prompt Composition}
\label{sec:prompting}

Given $\mathcal{P}^\text{shared}$ and $\mathcal{P}^\text{theme}$, we automatically generate detailed \emph{per-image} captions $\mathcal{P}^{non-shared}_i$ using a Vision-Language Model (VLM). The VLM is instructed to identify shared and non-shared elements, integrate the user-provided ideas while respecting the underlying structure, and describe per-image pose variations. More details in Supplementary Materials (SM). 

\subsection{\graph}
\label{sec:graph}

We define a graph $G = (V, E)$, where nodes $V = \{ I_i \}_{i=1}^N$ represent images, and edges $E = \{{i, j} \mid i \neq j,\ I_i, I_j \in V \}$ connect all distinct image pairs.
During the generation, each node is associated with a noisy latent $z_i^t$, and each edge with latent of a two-image grid $z^t_{ij}=\operatorname{concat}(z^t_i, z^t_j)$.
Edges are further assigned concatenated control depth maps and grid prompts:
$\mathcal{P}_{ij} = $
\textit{``Image grid of [$\mathcal{P}^\text{shared}$]. Left: [$\mathcal{P}_i^\text{non-shared}$]. Right: [$\mathcal{P}_j^\text{non-shared}$].''}

At each generation step, $\{z^t_{ij}\}_{e \in E}$ are constructed from $\{z^t_{i} \}_{i=1}^N$ and denoised jointly with \mff (\cref{sec:mff}) that encourages consistency across the entire graph by enforcing pre-computed 2D matches $M_{ij}$:
{
\setlength{\abovedisplayskip}{6pt}
\setlength{\belowdisplayskip}{6pt}
\begin{equation}
\scalebox{0.88}{$
\displaystyle
\{z^{t-1}_{ij}\}_{e \in E}=\operatorname{denoise}_{\operatorname{MFF}}(\{z^t_{ij}\}_{e \in E}, \{\mathcal{P}_{ij}\}_{e \in E}, \{M_{ij}\}_{e \in E})
$}
\label{eq:graph_denoise}
\end{equation}}
Since every node $z^t_i$ participates in multiple adjacent edges, a graph denoising step yields multiple versions of $z^{t-1}_{i | ij}$.
We thus consolidate pairwise latents $\{z^{t-1}_{ij} \}_{e \in E}$ back into image latents $\{z^{t-1}_{i} \}_{i=1}^N$ by extracting and averaging all $z^{t-1}_{i | ij}$, following \cite{bar2023multidiffusion}.
In practice, full graph connectivity is not required, as analyzed in \cref{sec:analysis}.

\begin{figure*}[ht]
  \centering
  \includegraphics[%
    width=1.0\linewidth,
    keepaspectratio
  ]{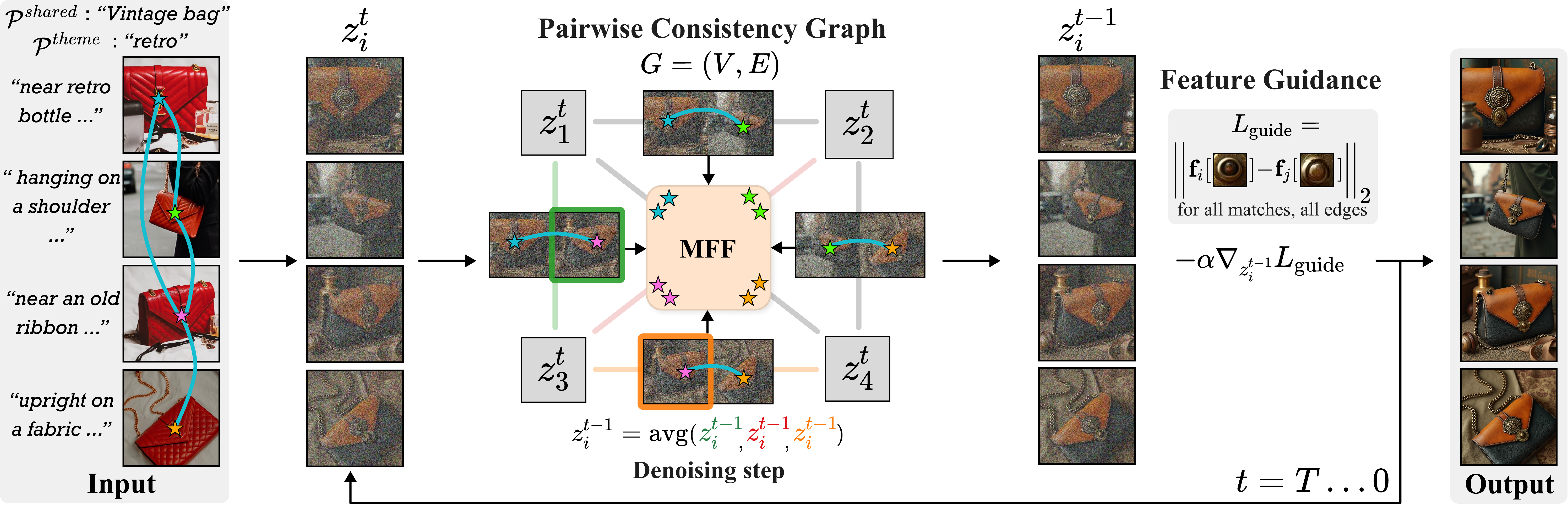}
  \vspace{-15pt}
  \caption{\textbf{\method pipeline.} Example for 4 images.  In pre-processing, pairwise matches are computed between all inputs, and per-image prompts are generated from the set-level prompts. At each denoising step, noisy image latents form a \graph, whose edges $z^{t}_{ij}$ are jointly denoised with \mff (MFF) and aggregated back into per-image latents $z^{t-1}_i$ by averaging over adjacent edges. The latents are further refined with \mguide via a feature-level matching objective.} \afterfigure
  \label{fig:method} 
\end{figure*}

\begin{figure}[t!]
  \centering
  \includegraphics[%
    width=1.0\linewidth,
    keepaspectratio
  ]{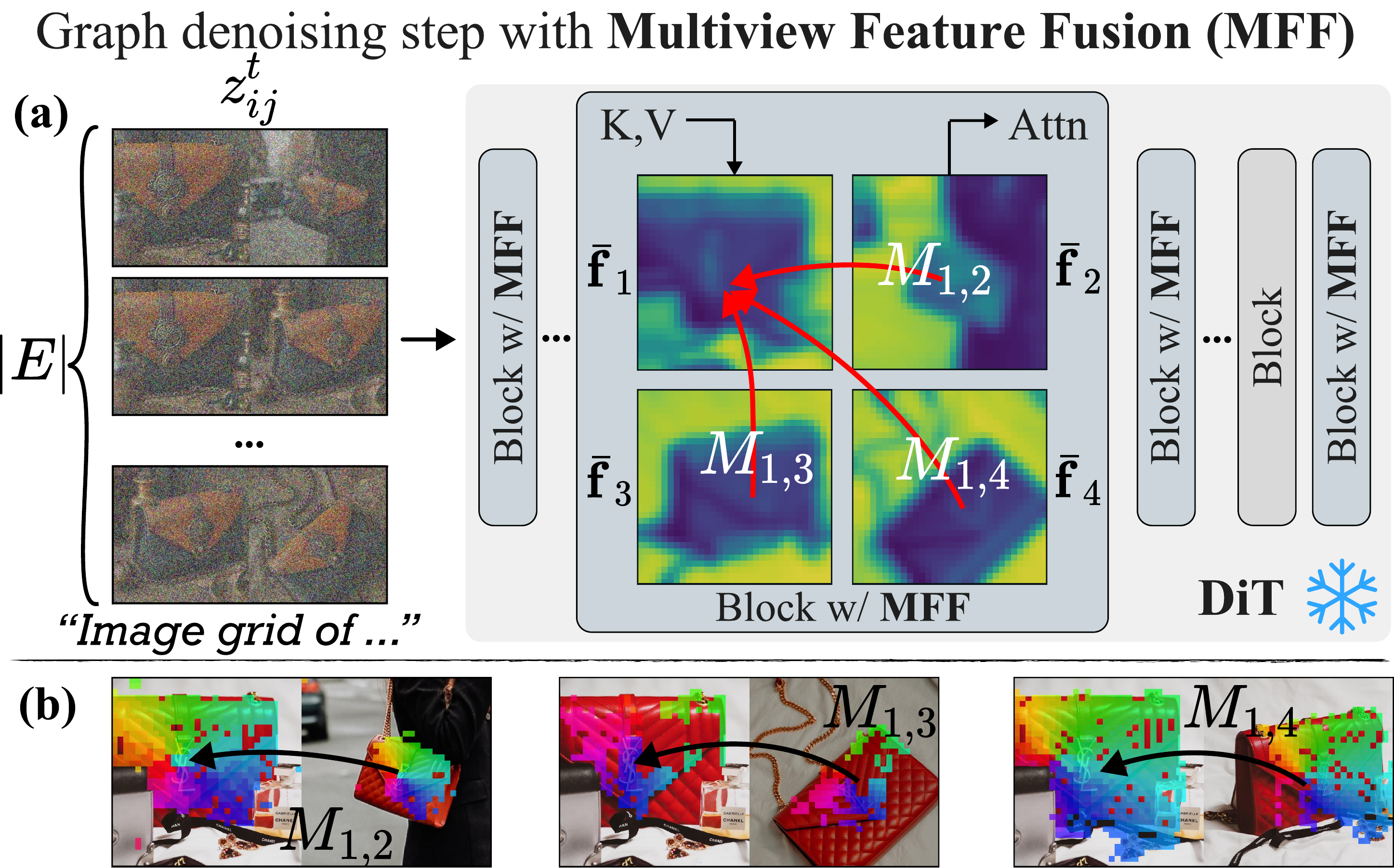}
  \vspace{-15pt}
  \caption{\textbf{MFF Denoising step.} (a) Two-image grids on all edges are denoised with a frozen DiT. Selected blocks average K,V along adjacent edges into $\bar{\mathbf{f}}_i$, which are then fused via source matches (b). Images are fused jointly, illustrated by arrows for $i{=}1$.} \afterfigure
  \label{fig:mff}
  \vspace{-3pt}
\end{figure}

\subsection{\mff}
\label{sec:mff}

The grid prior alone is insufficient for fine-grained alignment within an image pair (\cref{fig:grid_prior}). Moreover, denoising separate edges can yield noticeably different appearances, even when a shared image latent is used. To promote pairwise and global consistency, we follow prior work (\cref{sec:related_work}) and directly manipulate the model’s internal features.

\begin{figure}[t!]
  \centering
  \includegraphics[width=1.0\linewidth,keepaspectratio]{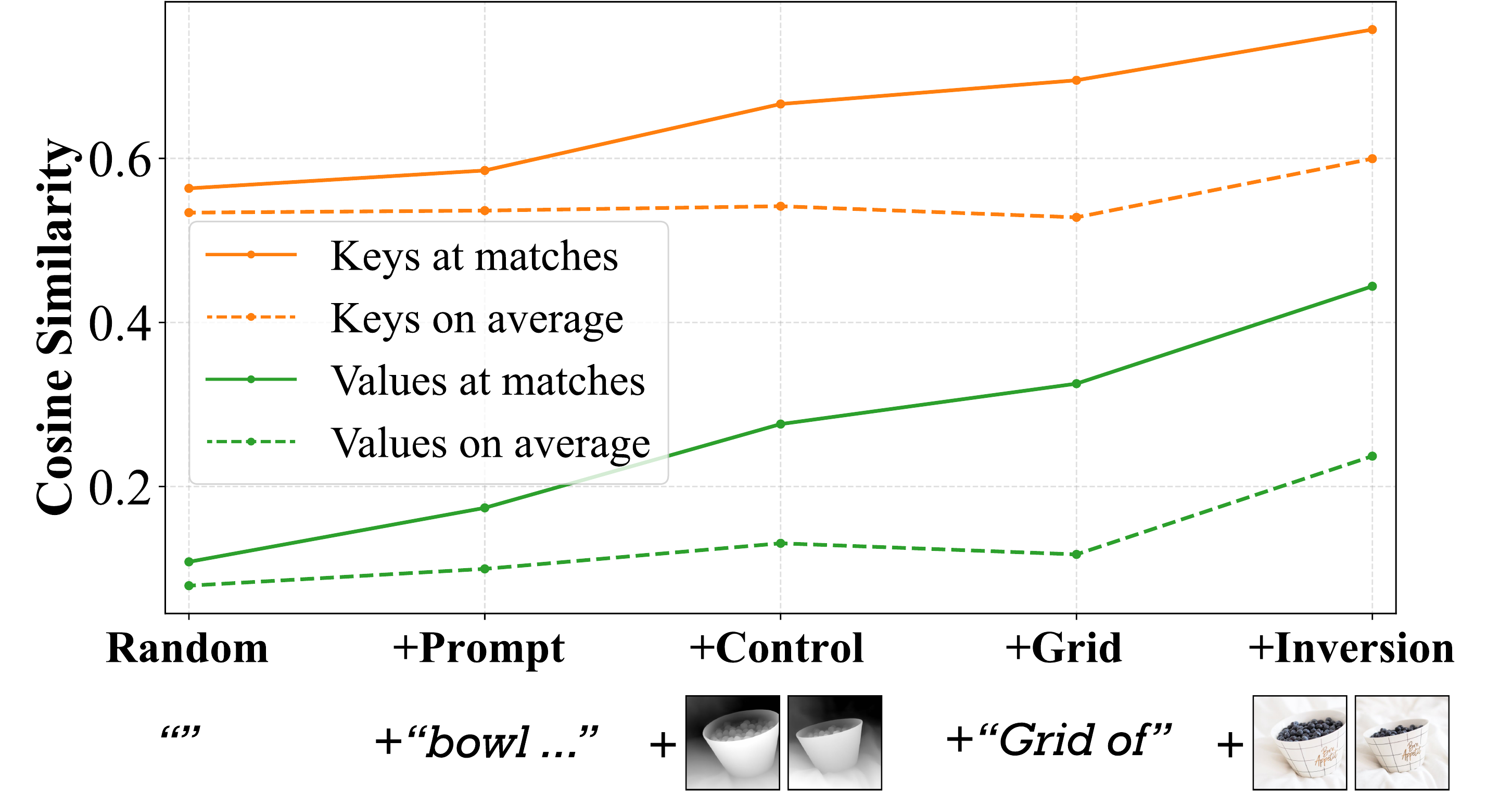}
  \vspace{-15pt}
  \caption{
  \textbf{Matched feature similarity vs. visual consistency.} 
  We consider increasingly consistent generation (left to right): random images $\to$ adding descriptive prompts $\to$ adding control signals $\to$ generating in a grid $\to$ DDIM \cite{song2020denoising} inversion which reconstructs fully consistent source images. 
  Keys and values differ in scale but follow the same pattern: cosine similarity at matched locations rises with consistency. Dashed lines show the baseline all-to-all similarity of feature maps. Points are averaged over 40 image pairs, all correspondences, blocks, and timesteps.} \afterfigure
  \label{fig:feat_sim}
\end{figure}

A natural choice for our task is to leverage off-the-shelf 2D correspondences extracted from the source images, which reliably capture shared content through confidence-based filtering.
We analyze the model’s feature space and observe that cosine similarity at these matched locations strongly correlates with generation consistency (\cref{fig:feat_sim}). Hence, we propose to promote it by increasing the similarity of matched features. 

Matches between each image pair are defined as a partial feature coordinate mapping 
$M_{ij}: \mathcal{C}_i \!\to\! \mathcal{C}_j$, where a coordinate $\mathbf{c} \in \mathcal{C}_i$ corresponds to $\mathbf{c}' = M_{ij}(\mathbf{c}) \in \mathcal{C}_j$ and $\mathcal{C}_i$ contains all matched points (\cref{fig:mff}b).
We first address pairwise consistency by considering a two-node graph with a single edge.
Let $\mathbf{f}_{ij} \in \mathbb{R}^{H\times 2W\times D} $ be a feature map from a model’s forward pass on a grid, interpreted as $\mathbf{f}_{ij}=\operatorname{concat}(\mathbf{f}_i,\mathbf{f}_j)$, with $\mathbf{f}_i,\mathbf{f}_j \in \mathbb{R}^{H\times W\times D}$. Given a pooling operator $\mathbf{f}[\mathbf{c}]$ extracting the vector at $\mathbf{c}$, 
\mff (MFF) fuses matched coordinates:
{
\setlength{\abovedisplayskip}{6pt}
\setlength{\belowdisplayskip}{6pt}
\begin{equation}
\label{eq:mff_grid}
\mathbf{f}_i[\mathbf{c}] \gets \tfrac{1}{2} \left(\mathbf{f}_i[\mathbf{c}] + \mathbf{f}_j[M_{ij}(\mathbf{c})]\right), \quad \forall \mathbf{c} \in \mathcal{C}_i.
\end{equation}
}
For an N-node graph, \cref{eq:mff_grid} is generalized by aggregating features of each image over its incident edges, followed by joint fusion across all images:
{
\setlength{\abovedisplayskip}{6pt}
\setlength{\belowdisplayskip}{6pt}
\begin{equation}
\bar{\mathbf{f}}_i = \frac{1}{|\delta(i)|}\!\sum_{e \in \delta(i)}\!\mathbf{f}_i^{e}, 
\quad \delta(i) = \{e \!\in\! E \mid i \!\in\! e\}
\end{equation}} \vspace{-0.2cm}
{
\begin{equation}
\label{eq:mff_graph}
\mathbf{f}_i[\mathbf{c}] \gets \tfrac{1}{N}\Bigl(
\bar{\mathbf{f}}_i[\mathbf{c}] + \sum_{j\ne i}\bar{\mathbf{f}}_j[M_{ij}(\mathbf{c})]\Bigr)
\end{equation}}
This stage is illustrated in \cref{fig:mff}. See SM for further details.

\subsection{\mguide}
\label{sec:guide}

\begin{figure*}[ht!]
\vspace{-0.9cm}
  \centering 
  \includegraphics[width=0.32\textwidth,keepaspectratio]{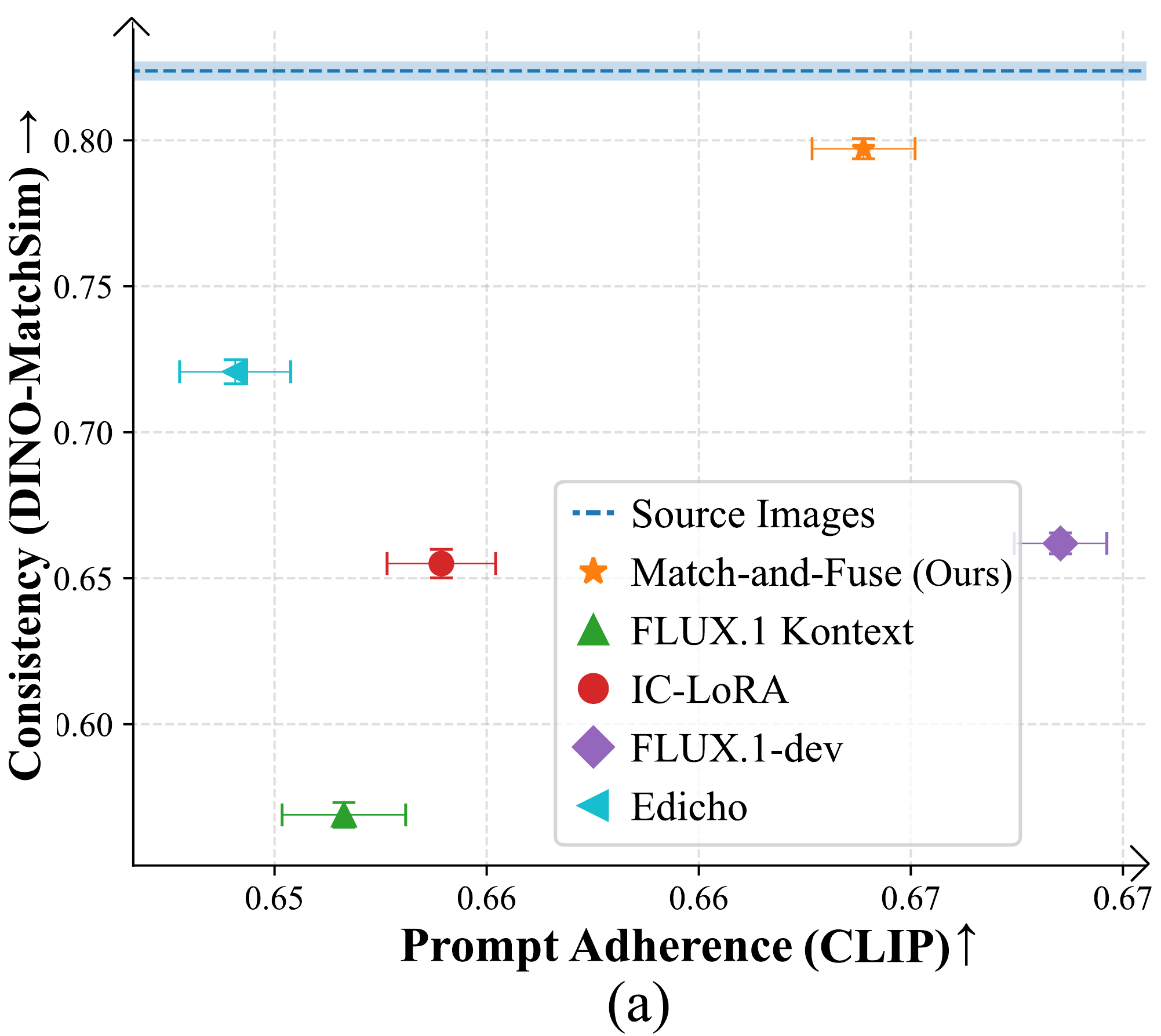} 
  \includegraphics[width=0.32\textwidth,keepaspectratio]{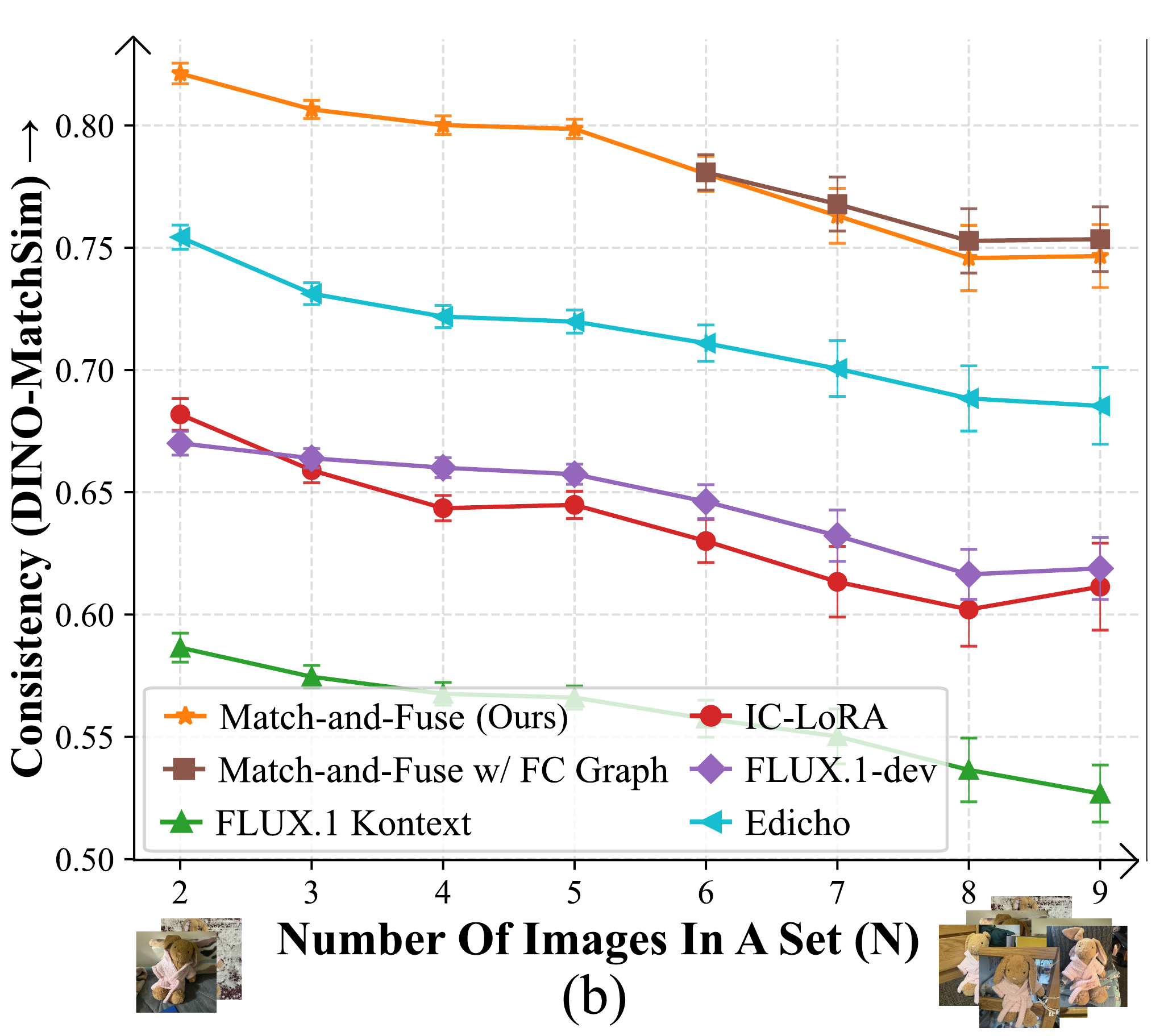} 
  \includegraphics[width=0.32\textwidth,keepaspectratio]{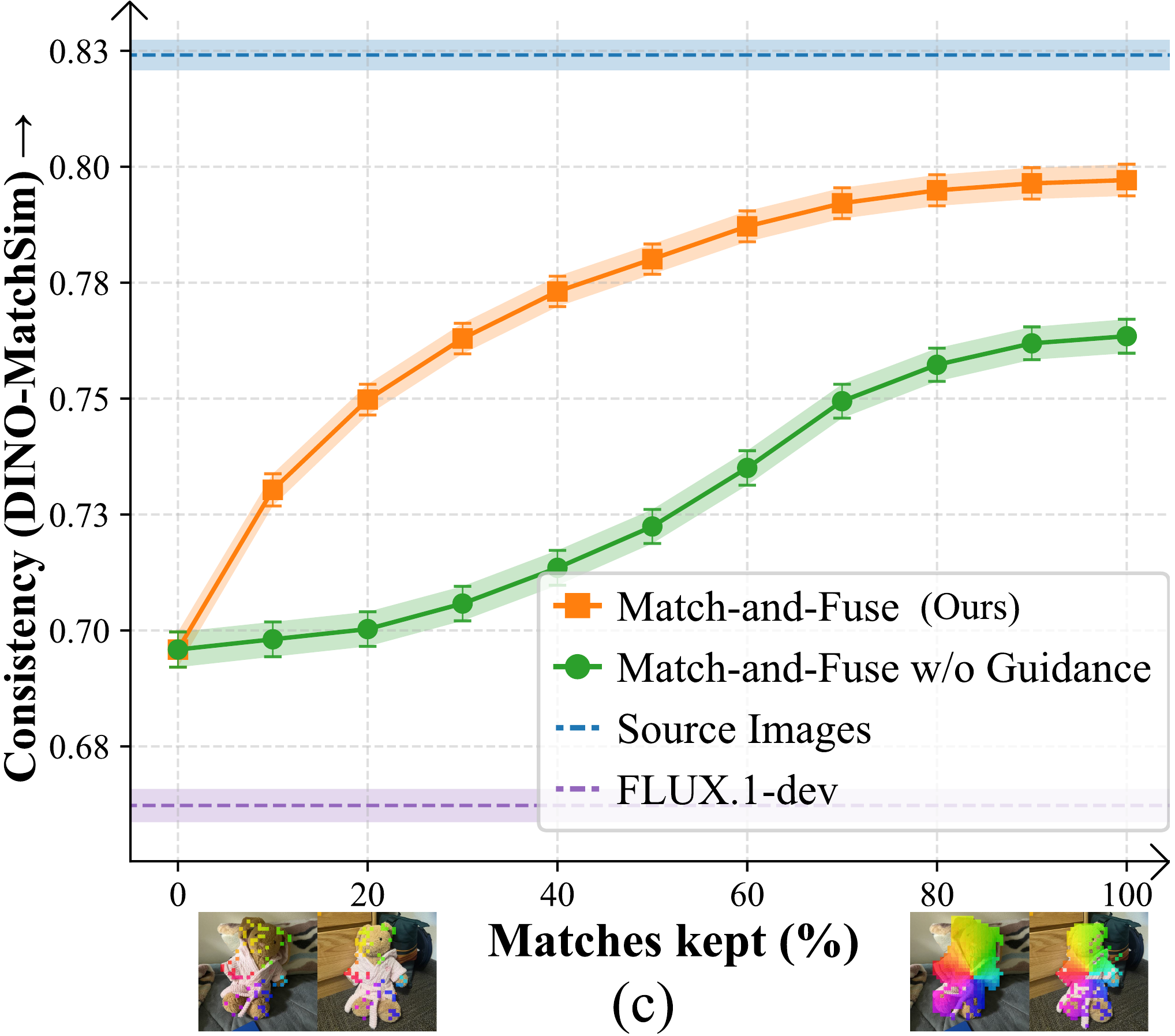}
  \vspace{-7pt} 
  \caption{\textbf{Quantitative comparison and analysis.} 
  We evaluate subject Consistency (DINO-MatchSim) of \method as a function (a) of Prompt Adherence (CLIP Score), compared to baselines; (b) Number of Images, compared to baselines and method variant with a fully-connected graph; (c) \% of Matches, compared to method variant w/o \mguide. 
  Error bars are SEM.} \afterfigure
  \label{fig:quantitative} 
\end{figure*}
To align small details, we further perform feature guidance \cite{epstein2023selfguidance, chefer2023attendandexciteattentionbasedsemanticguidance, shi2023dragdiffusion}, completing each generation step (\cref{fig:method}, right). 
Specifically, we define an optimization objective over all edges as the distance between matched features, and refine all node latents via gradient descent with $\nabla_{z^{t-1}_i}L_{\mathrm{guide}}$ of: 
{
\begin{equation}
\scalebox{0.90}{$
\displaystyle
L_{\mathrm{guide}} = \frac{1}{|E|}\sum_{\{i, j\} \in E}\frac{1}{|M_{ij}|}\sum_{\mathbf{c} \in M_{ij}}||\mathbf{f}_i[\mathbf{c}] - \mathbf{f}_j[M_{ij}(\mathbf{c})]||_2
$}
\end{equation}}
The feature maps $\mathbf{f}$ are extracted from an additional DiT forward pass.
MFF can be viewed as the analytical solution to this optimization problem operating during inference, while guidance corrects remaining inconsistencies directly in the latent space.
Using guidance as the primary driver would be expensive and risk pushing latents off-distribution, so we apply it only as a light refinement. The exact schedule is provided in SM. 
Because gradients propagate through the model, updates have a wider receptive field than the discrete match locations, improving robustness under sparse correspondences (see \cref{sec:analysis}).

\begin{figure}[t!]
  \centering
  \includegraphics[width=1.0 \linewidth,keepaspectratio]{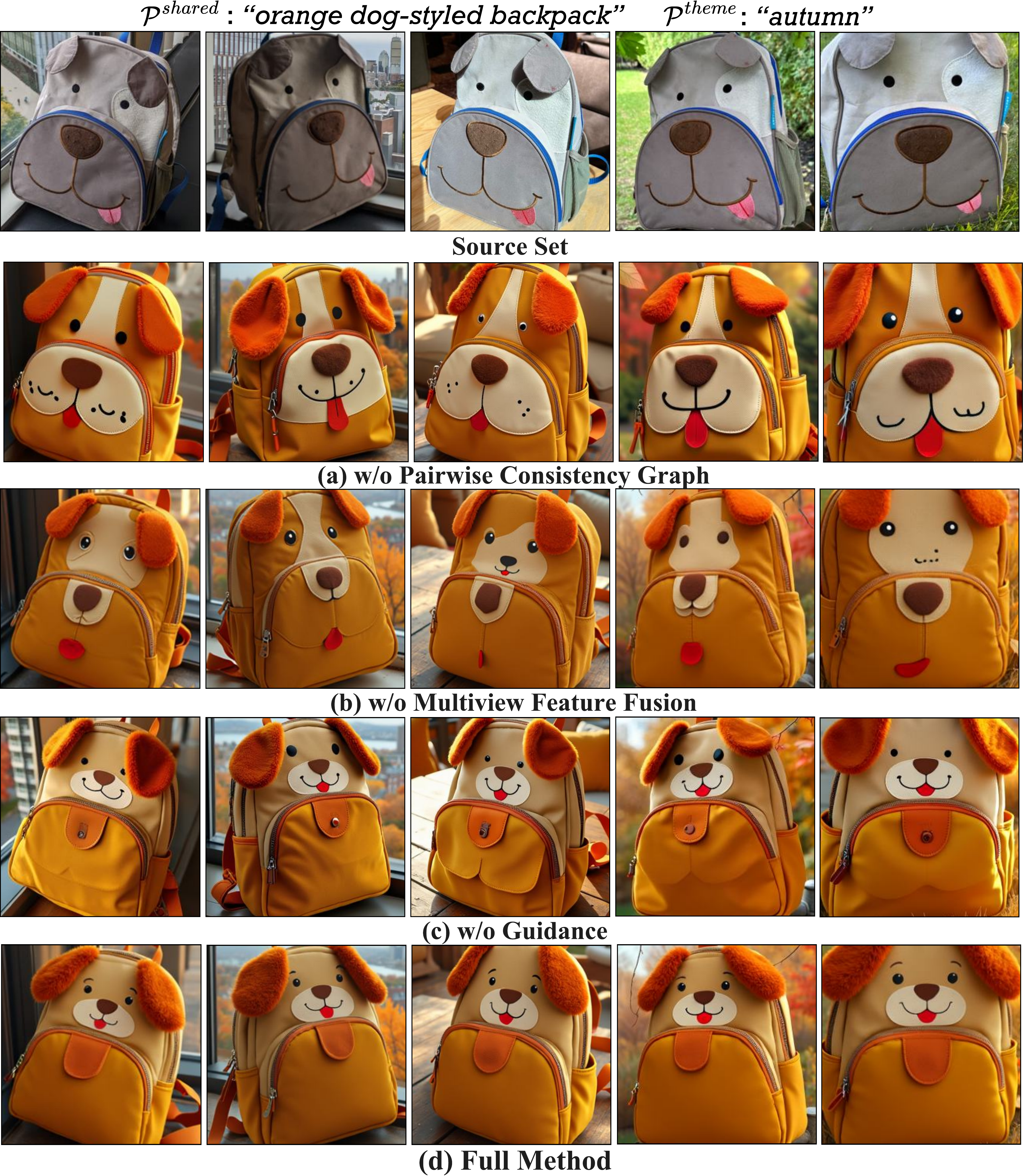}
  \vspace{-20pt}
  \caption{\textbf{Ablation.} Subtraction of each of our method's components (a-c) degrades consistency, demonstrating their necessity in the full method (d). Each setting is detailed in \cref{sec:analysis}.
  }
  \label{fig:ablation} \afterfigure
\end{figure}

\vspace{-1.5ex}
\section{Experiments}

\vspace{-1.5ex}
\setlength{\tabcolsep}{2pt}
\begin{table}
    \centering
    \scalebox{0.88}{
    \begin{tabular}{lccc}
        \toprule
                               &    CLIP $\uparrow$   &  DreamSim $\uparrow$  &  DINO-MatchSim $\uparrow$  \\
        \midrule
        FLUX Kontext         &     0.65 $\pm$ 0.002  &  0.78 $\pm$ 0.004  &     0.57 $\pm$ 0.004  \\
        IC-LoRA                &     0.65 $\pm$ 0.001  &  0.71 $\pm$ 0.006  &     0.65 $\pm$ 0.004  \\
        FLUX             &  \textbf{0.67 $\pm$ 0.001}  &  0.76 $\pm$ 0.004  &     0.66 $\pm$ 0.004  \\
        Edicho                 &     0.65 $\pm$ 0.001  &  0.81 $\pm$ 0.004         &     0.72 $\pm$ 0.004  \\
        \midrule
        \method                &     0.66 $\pm$ 0.001  &  \textbf{0.85 $\pm$ 0.004}  &  \textbf{0.80 $\pm$ 0.003}  \\
        \midrule
        w/o Guidance           &     0.66 $\pm$ 0.001  &  0.82 $\pm$ 0.004  &     0.76 $\pm$ 0.004  \\
        w/o MFF                &     0.66 $\pm$ 0.001  &  0.83 $\pm$ 0.004  &     0.78 $\pm$ 0.003  \\
        w/o Pairwise Graph              &     0.66 $\pm$ 0.001  &  0.82 $\pm$ 0.004  &     0.75 $\pm$ 0.004  \\
        \bottomrule
    \end{tabular}
    } \vspace{-0.2cm}
    \caption{\textbf{Quantitative evaluation.} We compare our method to baselines and ablate method components.}
    \label{tab:num_eval}
    \vspace{-12pt}
    
\end{table}

\begin{figure*}[!htbp]
  \centering
    \includegraphics[width=1.\textwidth]{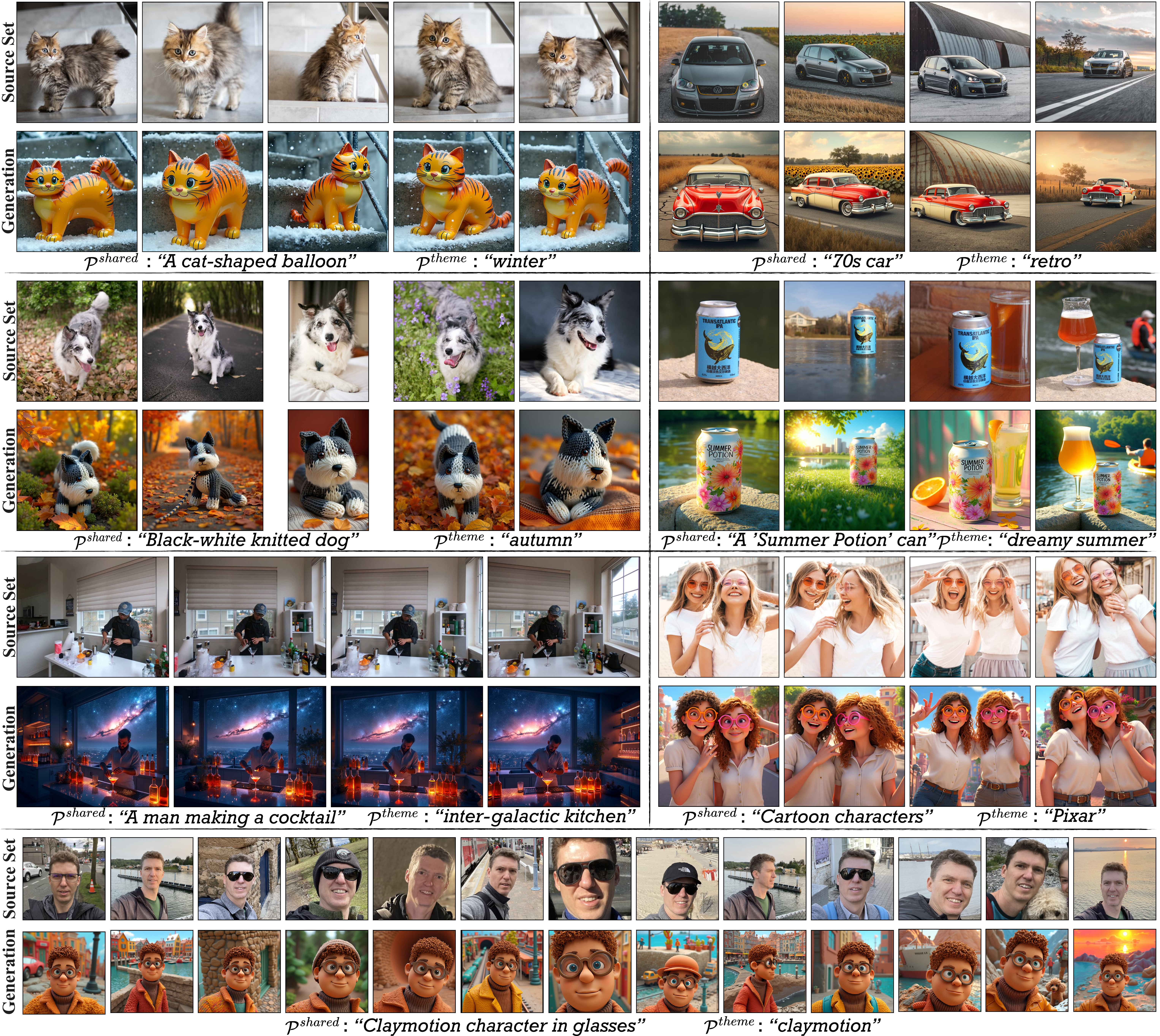}
    \vspace{-12pt}
  \caption{\textbf{Qualitative results.} 
  \method generates consistent content of rigid and non-rigid shared elements, single and multi-subject, with shared or varying background, preserving fine-grained consistency in textures, small details, and typography. Notably, it can generate consistent long sequences (last row). See SM for full sets of results.} \afterfigure
  \label{fig:results}
\end{figure*}

\begin{figure*}[ht]
  \centering
  \includegraphics[width=1.0 \textwidth]{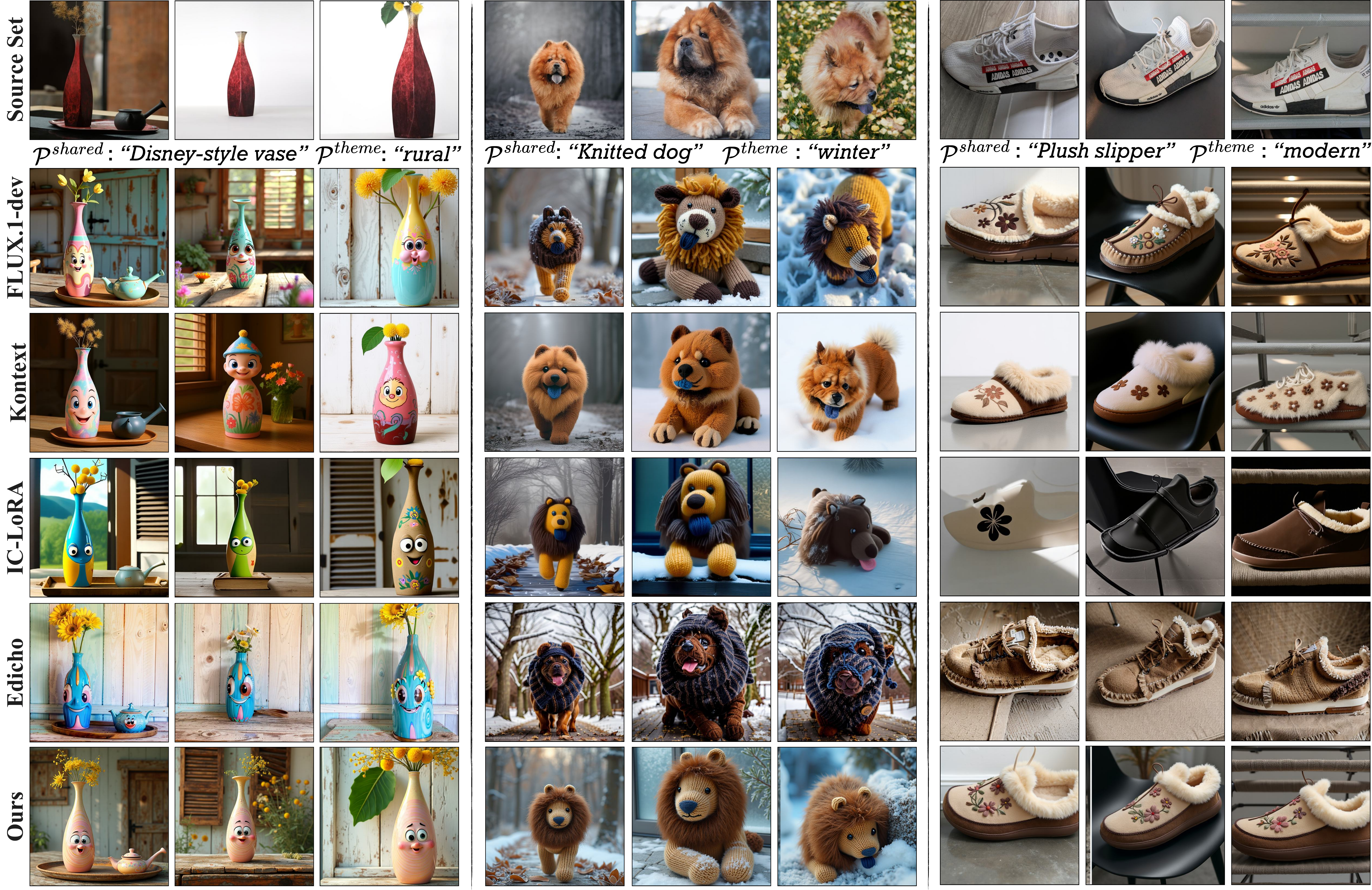}
  \vspace{-15pt}
  \caption{\textbf{Qualitative comparisons.}  \method (Ours) generates image sets with the highest consistency compared to our baselines FLUX \cite{flux2024}, FLUX Kontext \cite{labs2025flux1kontextflowmatching}, IC-LoRA \cite{lhhuang2024iclora}, and Edicho \cite{bai2024edicho}. See \cref{sec:compare} for comments on each result and SM for full image sets.} \afterfigure
  \label{fig:qualitative}
\end{figure*}

\begin{figure}[ht]
  \centering
  \includegraphics[width=\linewidth,keepaspectratio]{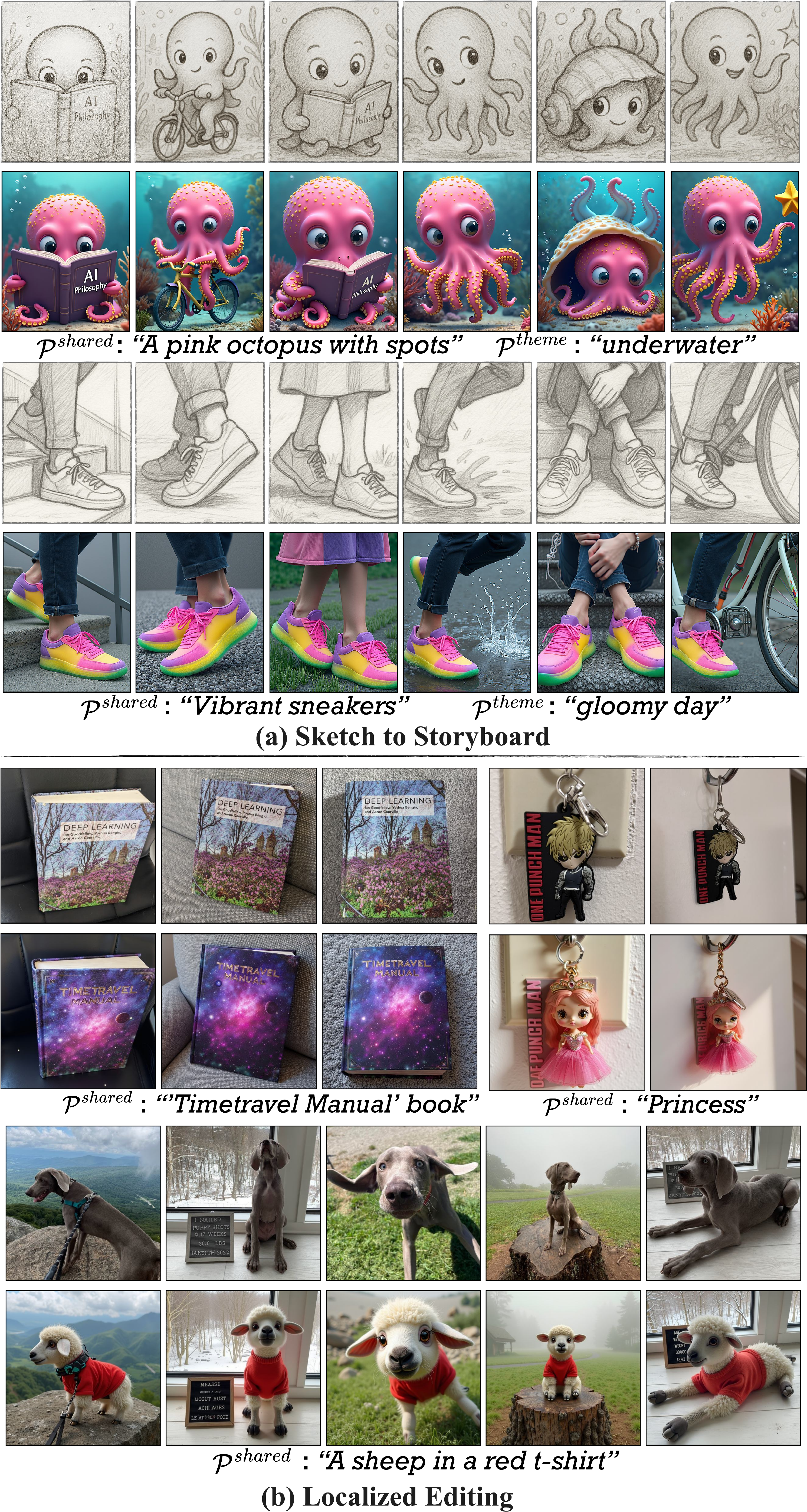}
  \vspace{-15pt}
  \caption{\textbf{Extended Applications.} Our method generalizes across various settings, enabling (a) consistent story generation from sketches and (b) localized editing omitting $\mathcal{P}^{theme}$ via FlowEdit \cite{kulikov2024flowedit} integration. See full sets of results in SM.
  } \afterfigure
  \label{fig:apps}
\end{figure}

We extensively evaluate \method.
Our implementation builds on FLUX \cite{flux2024} with depth-conditioning \cite{fluxcontrolnet2024}, and uses RoMA \cite{edstedt2023romarobustdensefeature} for matching. Full details are in SM.

\vspace{-1.2ex}
\subsection{Evaluation Setup}
\label{sec:setup}
Evaluating our task requires image sets that share content across diverse scenarios, along with a metric that captures fine-grained cross-image consistency. Since no existing benchmark or metric supports this setting, we introduce a setup tailored to consistent set-to-set generation.

\myparagraph{{\bf Evaluation set.}}
We curate a benchmark with 400 edits for 149 diverse distinct image sets, 3–15 images in each, combining all sets from \cite{kumari2022customdiffusion} and \cite{ruiz2023dreambooth} with frames sampled from 3D datasets \cite{instructnerf2023,barron2021mipnerf,li2022neural3dvideosynthesis}, keyframes of a few publicly sourced videos, and sketched storyboards generated by ChatGPT. Data will be publicly released. 

\begin{table}[t]
\centering
\setlength{\tabcolsep}{2pt}

\begin{minipage}{0.48\linewidth}
\centering
\scalebox{0.88}{
\begin{tabular}{lcc}
    \toprule
                      & Users $\uparrow$ & VLM $\uparrow$ \\
    \midrule
    Kontext           & 88\%  & 82\% \\
    IC-LoRA           & 90\%  & 92\% \\
    FLUX              & 92\%  & 94\% \\
    Edicho            & 83\%  & 78\% \\
    \bottomrule
\end{tabular}
}
\\[2pt]
\small (a)
\end{minipage}
\hfill
\begin{minipage}{0.48\linewidth}
\centering
\scalebox{0.88}{
\begin{tabular}{lc}
    \toprule
                   & Agreement $\uparrow$ \\
    \midrule
    DreamSim        & 84.3 \\
    VLM             & 84.9 \\
    DINO-MatchSim   & \textbf{91.4} \\
    \bottomrule
\end{tabular}
}
\\[2pt]
\small (b)
\end{minipage}

\vspace{-0.15cm}
\caption{
(a) \textbf{User Study and VLM evaluation.} 2AFC voting demonstrates the preference of humans and VLM of our method over all baselines.
(b) \textbf{Agreement between metrics with human judgments.} \% of samples for which the winner according to a higher metric agrees with the users' majority vote. Average across comparisons and baselines.
}
\label{tab:win_rate} \afterfigure
\end{table}

\myparagraph{{\bf Metrics.}} We quantitatively evaluate our method, beginning with fine-grained cross-image consistency.
Previous work relies on \emph{global} visual-similarity metrics to assess set consistency—for example, \cite{tewel2024training} uses DreamSim \cite{fu2023dreamsim} on object-masked regions. However, as our user study (\cref{tab:win_rate}b) shows, such global metrics are not ideal. 
To address this, we introduce \textit{DINO-MatchSim}. For each image pair, we compute patch-level nearest-neighbor correspondences $\textit{NN}_{ij}$ between DINOv3 \cite{simeoni2025dinov3} feature maps from the source images, and measure similarity at the corresponding output locations:
{
\setlength{\abovedisplayskip}{6pt}
\setlength{\belowdisplayskip}{6pt}
\begin{equation}
\textit{S}_{ij} = \frac{1}{|\textit{NN}_{ij}|} \sum_{(p,q) \in \textit{NN}_{ij}} \cos\big(\tilde{F}_i(p), \tilde{F}_j(q)\big),
\label{eq:dino_matchsim}
\end{equation}}
where $\tilde{F}_i$ denotes the $i^{\text{th}}$ output feature map. The average similarity across all image pairs is \textit{DINO-MatchSim}.
For prompt adherence, we report the average CLIP score over the set. To assess human preference, we conduct a large-scale user study on a random subset of the benchmark, comparing \method to a baseline in a randomized two-alternative forced-choice (2AFC) format. Finally, we include a VLM-based evaluation using GPT-5 \cite{openai2025gpt5}, which received the same content as the human raters. 

\vspace{-5pt}
\subsection{Qualitative results}
\label{sec:qualitative}
Sample qualitative results of our method are shown in \cref{fig:results,fig:teaser,fig:qualitative}, demonstrating that it performs robustly under non-rigid pose variations, diverse viewpoints, and partial occlusions, on single or multi-object sets (\emph{e.g.} two girls). The output sets maintain fine-grained consistency in textures (\emph{e.g.} crochet, balloon and can print), small elements (\emph{e.g.} kitchen items), and typography (\emph{e.g.} the can title). Notably, \method is able to handle big number of images (\emph{e.g.} $N=13$ in \cref{fig:results}, last row). 

\vspace{-1.5ex}
\subsection{Comparisons}
\label{sec:compare}

Since no existing method is designed for our set-to-set generation task, we compare \method to the closest baselines:
\textbf{(1)} FLUX \cite{flux2024} with ControlNet \cite{fluxcontrolnet2024} conditioning serves as a baseline for assessing independent generation using the same base model as our method. \textbf{(2)} FLUX Kontext \cite{labs2025flux1kontextflowmatching},
an image-to-image editing model. 
While it provides consistent edits over multiple turns on a single image, it lacks an explicit mechanism for achieving consistency across a single-turn image set editing.
\textbf{(3)} IC-LoRA \cite{lhhuang2024iclora}, 
text-driven generation of image sets with shared elements by fine-tuning LoRA modules for specific context types. We use their most relevant public checkpoint and add spatial ControlNet conditioning for comparability. 
\textbf{(4)} Edicho \cite{bai2024edicho}, the closest baseline to our method, combines depth-conditioned generation with explicit correspondences, but is limited to an image pair editing through a single reference.
Additional implementation details are in \cref{sec:impl_base}. 

\Cref{fig:qualitative} shows representative qualitative results.
FLUX \cite{flux2024} produces inconsistencies in both coarse and fine details.
FLUX Kontext \cite{labs2025flux1kontextflowmatching} has low prompt adherence (\emph{e.g.}, a toy remains animal-like), and often distorts object structure (\emph{e.g.} vase, dog).
IC-LoRA \cite{lhhuang2024iclora} achieves partial consistency, with coherence often restricted to subsets (vase). Its realism and fidelity are notably lower than FLUX. Edicho \cite{bai2024edicho} performs best among baselines but still shows noticeable inconsistencies (dog), as its one-to-all warping enforces consistency only with the first image, making the choice of an anchor ambiguous and degrading coherence across views.
In contrast, \method (bottom row) maintains high image quality while substantially improving structural and fine-grained consistency. We additionally compare to the closed-source Nano Banana API \cite{google2025nanobanana} and find that it does not solve our task (SM).

We quantitatively evaluate all methods using the metrics in \cref{sec:setup}. \Cref{fig:quantitative}a plots Consistency (DINO-MatchSim) versus Prompt adherence (CLIP score).
As an upper bound, we report DINO-MatchSim on the source sets, which are consistent by definition.
Our method achieves the highest consistency among baselines, approaching the source score, while maintaining a comparable CLIP score.
Since DINO-MatchSim evaluates consistency relative to the source structure, methods that distort shape, such as FLUX Kontext, score lowest. \cref{tab:num_eval} further shows that \method attains the highest DreamSim score.
We complement these results with a user study and VLM-based evaluation. As shown in \cref{tab:win_rate}a, participants and VLM preferred our method over all baselines.

\subsection{Analysis \& Ablations}
\label{sec:analysis}
\myparagraph{{\bf Ablation.}} 
We ablate each component in \cref{fig:ablation} and \cref{tab:num_eval}.
\textit{W/o \graph}, pairwise steps become single-image predictions, with MFF and Guidance applied per image. Correspondences alone cannot align appearances, leading to identity drift.
In \textit{w/o \mff}, pairwise updates use only the grid prior; latent versions aggregated across edges diverge more easily, reducing consistency.
\textit{Omitting \mguide} at each step leads to misaligned fine-grained details.

\myparagraph{{\bf Number of images.}}
A fully connected graph has $O(N^2)$ edges. To maintain scalability, we limit each node degree to 4 random neighbors, yielding full connectivity for $N \leq 5$ and increasing sparsity thereafter. This keeps runtime linear while preserving cross-image communication. We use this setup in all our experiments.
\Cref{fig:quantitative}b shows consistency as a function of $N$. For each $N$-image set, we evaluate subsets of the first $N' \in [2, N]$ images and report averages over all $N'$. Our method is slightly below its fully connected variant but remains similar in performance. Although \method degrades with $N$ it remains more consistent for 9 images than baselines do for 2.

\myparagraph{{\bf Match sparsity.}}
We evaluate robustness to reduced correspondences by randomly subsampling x\% of matches. As shown in \cref{fig:quantitative}c, DINO-MatchSim remains high even under strong sparsification, whereas the variant without \mguide drops more rapidly—highlighting its role in maintaining consistency with sparse matches.

\begin{figure}[t!]
  \centering
  \includegraphics[width=1.0\linewidth]{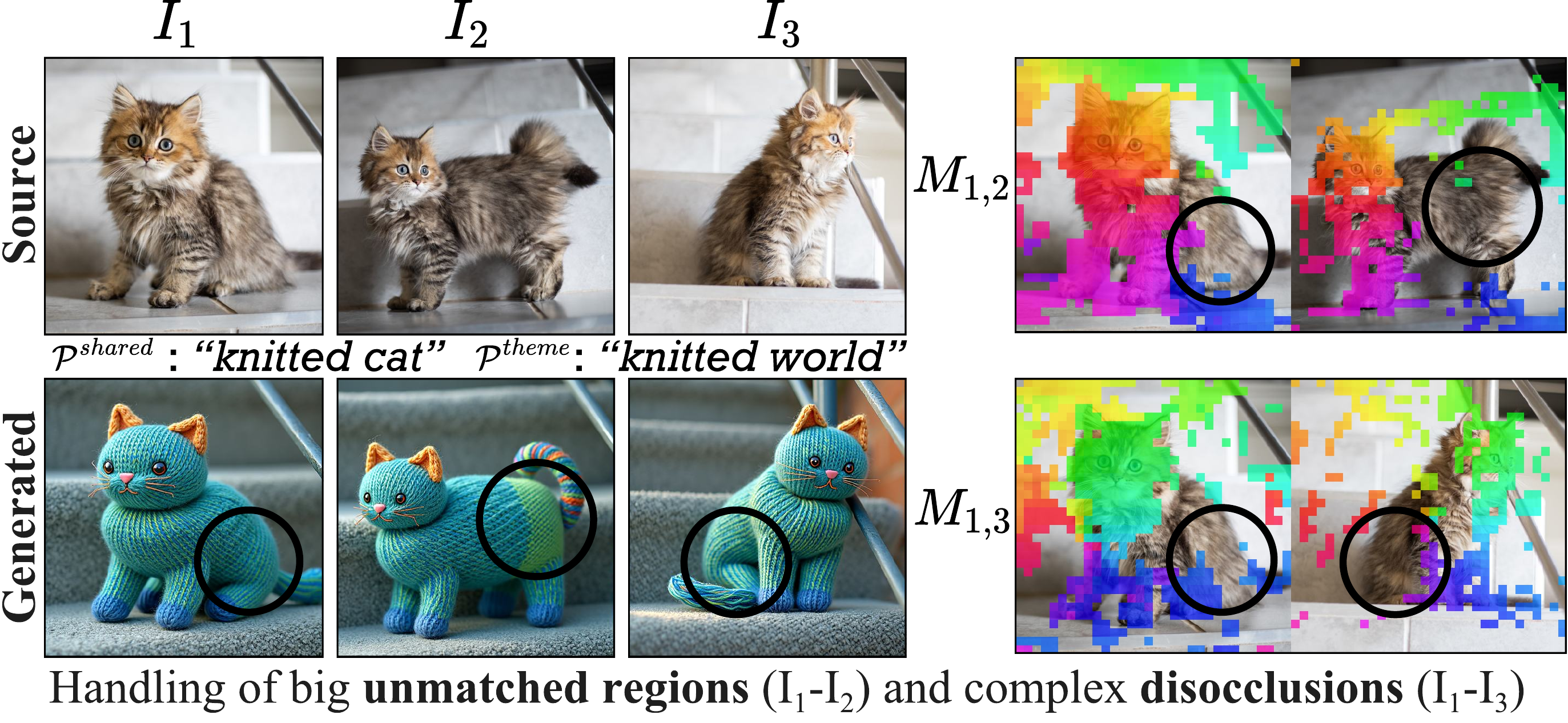}
  \vspace{-20pt}
  \caption{\textbf{Limitations.} 
  The method may produce inconsistencies in largely unmatched regions ($I_1$, $I_2$) or under complex disocclusions (symmetry in $I_2$, $I_3$). 
  } \afterfigure
  \label{fig:limitations}
\end{figure}

\subsection{Extended Applications}
\label{sec:apps}
As shown in \cref{fig:apps}a, our method generalizes to sketched source sets, enabling consistent visualization of \textit{storyboards}. 
In \cref{fig:apps}b, we further demonstrate \textit{localized editing} by combining our method with FlowEdit \cite{kulikov2024flowedit}. We provide default integration parameters in the SM; however, as is a known limitation of FlowEdit, achieving the desired balance between structure preservation and appearance change often requires mild per-edit hyperparameter tuning. Automating this selection is left for future work.

\section{Discussion and Conclusions}

We introduced \method{}, the first method for controlled generation from unstructured image collections. Extensive experiments demonstrate that it outperforms strong baselines and enables diverse creative applications, advancing a largely underexplored yet fundamental visual modality. Several limitations remain for future work. Our performance depends on the quality and density of pixel correspondences, which may result in inconsistencies in largely unmatched or ambiguous regions (e.g., disocclusions or symmetries; see \cref{fig:limitations}). It also relies on the ability of the base model to preserve the source conditioning depth maps, which may occasionally deviate from the source layout, depending on the target prompt and generative prior of the T2I model (\emph{e.g.} body and hand poses of the girls in \cref{fig:results}). We believe these insights lay a path toward future extensions, such as video collections and foundation models for set-to-set generation.

\section*{Acknowledgments}
We thank Danah Yatim for her valuable comments. We thank Vladimir Kulikov for insightful discussions on FlowEdit. We thank Qingyan Bai and the other authors of Edicho for conducting qualitative comparisons with their method.


{
    \small
    \bibliographystyle{ieeenat_fullname}
    \bibliography{main}

@String(CVPR= {IEEE Conf. Comput. Vis. Pattern Recog.})

@String(ICCV= {Int. Conf. Comput. Vis.})

@String(ECCV= {Eur. Conf. Comput. Vis.})

@String(TOG= {ACM Trans. Graph.})

@String(ICLR = {Int. Conf. Learn. Represent.})

@String(CVPR  = {CVPR})

@String(ICCV  = {ICCV})

@String(ECCV  = {ECCV})

@String(TOG   = {ACM TOG})

@String(ICLR  = {ICLR})

@misc{flux2024,
  author       = {{Black-Forest}},
  title        = {Flux: Diffusion models for layered image generation},
  howpublished = {\url{https://github.com/black-forest-labs/flux}},
  year         = {2024},
  note         = {Accessed: 2024-09-24}
}

@misc{fluxcontrolnet2024,
  author       = {{XLabs-AI}},
  title        = {Flux-ControlNet Collections},
  howpublished = {\url{https://huggingface.co/XLabs-AI/flux-controlnet-collections}},
  year         = {2024},
  note         = {Accessed: 2024-11-25}
}

@misc{zhang2023addingconditionalcontroltexttoimage,
      title={Adding Conditional Control to Text-to-Image Diffusion Models}, 
      author={Lvmin Zhang and Anyi Rao and Maneesh Agrawala},
      year={2023},
      eprint={2302.05543},
      archivePrefix={arXiv},
      primaryClass={cs.CV},
      url={https://arxiv.org/abs/2302.05543}, 
}

@misc{yang2024depthv2,
      title={Depth Anything V2}, 
      author={Lihe Yang and Bingyi Kang and Zilong Huang and Zhen Zhao and Xiaogang Xu and Jiashi Feng and Hengshuang Zhao},
      year={2024},
      eprint={2406.09414},
      archivePrefix={arXiv},
      primaryClass={cs.CV},
      url={https://arxiv.org/abs/2406.09414}, 
}

@misc{xie2015holisticallynestededgedetection,
      title={Holistically-Nested Edge Detection}, 
      author={Saining Xie and Zhuowen Tu},
      year={2015},
      eprint={1504.06375},
      archivePrefix={arXiv},
      primaryClass={cs.CV},
      url={https://arxiv.org/abs/1504.06375}, 
}

@misc{shi2020improvingimagecaptioningbetter,
      title={Improving Image Captioning with Better Use of Captions}, 
      author={Zhan Shi and Xu Zhou and Xipeng Qiu and Xiaodan Zhu},
      year={2020},
      eprint={2006.11807},
      archivePrefix={arXiv},
      primaryClass={cs.CV},
      url={https://arxiv.org/abs/2006.11807}, 
}

@misc{saharia2022photorealistictexttoimagediffusionmodels,
      title={Photorealistic Text-to-Image Diffusion Models with Deep Language Understanding}, 
      author={Chitwan Saharia and William Chan and Saurabh Saxena and Lala Li and Jay Whang and Emily Denton and Seyed Kamyar Seyed Ghasemipour and Burcu Karagol Ayan and S. Sara Mahdavi and Rapha Gontijo Lopes and Tim Salimans and Jonathan Ho and David J Fleet and Mohammad Norouzi},
      year={2022},
      eprint={2205.11487},
      archivePrefix={arXiv},
      primaryClass={cs.CV},
      url={https://arxiv.org/abs/2205.11487}, 
}

@misc{rombach2022highresolutionimagesynthesislatent,
      title={High-Resolution Image Synthesis with Latent Diffusion Models}, 
      author={Robin Rombach and Andreas Blattmann and Dominik Lorenz and Patrick Esser and Björn Ommer},
      year={2022},
      eprint={2112.10752},
      archivePrefix={arXiv},
      primaryClass={cs.CV},
      url={https://arxiv.org/abs/2112.10752}, 
}

@misc{esser2024scalingrectifiedflowtransformers,
      title={Scaling Rectified Flow Transformers for High-Resolution Image Synthesis}, 
      author={Patrick Esser and Sumith Kulal and Andreas Blattmann and Rahim Entezari and Jonas Müller and Harry Saini and Yam Levi and Dominik Lorenz and Axel Sauer and Frederic Boesel and Dustin Podell and Tim Dockhorn and Zion English and Kyle Lacey and Alex Goodwin and Yannik Marek and Robin Rombach},
      year={2024},
      eprint={2403.03206},
      archivePrefix={arXiv},
      primaryClass={cs.CV},
      url={https://arxiv.org/abs/2403.03206}, 
}

@misc{meng2022sdeditguidedimagesynthesis,
      title={SDEdit: Guided Image Synthesis and Editing with Stochastic Differential Equations}, 
      author={Chenlin Meng and Yutong He and Yang Song and Jiaming Song and Jiajun Wu and Jun-Yan Zhu and Stefano Ermon},
      year={2022},
      eprint={2108.01073},
      archivePrefix={arXiv},
      primaryClass={cs.CV},
      url={https://arxiv.org/abs/2108.01073}, 
}

@article{kulikov2024flowedit,
	title = {FlowEdit: Inversion-Free Text-Based Editing Using Pre-Trained Flow Models},
	author = {Kulikov, Vladimir and Kleiner, Matan and Huberman-Spiegelglas, Inbar and Michaeli, Tomer},
	journal = {arXiv preprint arXiv:2412.08629},
	year = {2024}
	}

@article{tewel2024training,
  title={Training-free consistent text-to-image generation},
  author={Tewel, Yoad and Kaduri, Omri and Gal, Rinon and Kasten, Yoni and Wolf, Lior and Chechik, Gal and Atzmon, Yuval},
  journal={ACM Transactions on Graphics (TOG)},
  volume={43},
  number={4},
  pages={1--18},
  year={2024},
  publisher={ACM New York, NY, USA}
}

@inproceedings{bai2024edicho,
  title     = {Edicho: Consistent Image Editing in the Wild},
  author    = {Bai, Qingyan and Ouyang, Hao and Xu, Yinghao and Wang, Qiuyu and Yang, Ceyuan and Cheng, Ka Leong and Shen, Yujun and Chen, Qifeng},
  booktitle = {arXiv preprint arXiv:2412.21079},
  year      = {2024}
}

@article{lhhuang2024iclora,
  title={In-Context LoRA for Diffusion Transformers},
  author={Huang, Lianghua and Wang, Wei and Wu, Zhi-Fan and Shi, Yupeng and Dou, Huanzhang and Liang, Chen and Feng, Yutong and Liu, Yu and Zhou, Jingren},
  booktitle={arXiv preprint arxiv:2410.23775},
  year={2024}
}

@article{tokenflow2023,
    title = {TokenFlow: Consistent Diffusion Features for Consistent Video Editing},
    author = {Geyer, Michal and Bar-Tal, Omer and Bagon, Shai and Dekel, Tali},
    journal={arXiv preprint arxiv:2307.10373},
    year={2023}
    }

@misc{garibi2025tokenverseversatilemulticonceptpersonalization,
title={TokenVerse: Versatile Multi-concept Personalization in Token Modulation Space}, 
author={Daniel Garibi and Shahar Yadin and Roni Paiss and Omer Tov and Shiran Zada and Ariel Ephrat and Tomer Michaeli and Inbar Mosseri and Tali Dekel},
year={2025},
eprint={2501.12224},
archivePrefix={arXiv},
primaryClass={cs.CV},
url={https://arxiv.org/abs/2501.12224}, 
}

@article{wang2024cove,
    title={COVE: Unleashing the Diffusion Feature Correspondence for Consistent Video Editing},
    author={Wang, Jiangshan and Ma, Yue and Guo, Jiayi and Xiao, Yicheng and Huang, Gao and Li, Xiu},
    journal={arXiv preprint arXiv:2406.08850},
    year={2024}
  }

@inproceedings{instructnerf2023,
         author = {Haque, Ayaan and Tancik, Matthew and Efros, Alexei and Holynski, Aleksander and Kanazawa, Angjoo},
         title = {Instruct-NeRF2NeRF: Editing 3D Scenes with Instructions},
         booktitle = {Proceedings of the IEEE/CVF International Conference on Computer Vision},
         year = {2023},
        }

@inproceedings{vicanerf2023,
      author = {Dong, Jiahua and Wang, Yu-Xiong},
      title = {ViCA-NeRF: View-Consistency-Aware 3D Editing of Neural Radiance Fields},
      booktitle = {NeurIPS},
      year = {2023},
     }

@article{wang2024view,
      title={View-Consistent 3D Editing with Gaussian Splatting},
      author={Wang, Yuxuan and Yi, Xuanyu and Wu, Zike and Zhao, Na and Chen, Long and Zhang, Hanwang},
      journal={arXiv preprint arXiv:2403.11868},
      year={2024}
  }

@article{gaussctrl2024,
author = {Wu, Jing and Bian, Jia-Wang and Li, Xinghui and Wang, Guangrun and Reid, Ian and Torr, Philip and Prisacariu, Victor},
title = {{GaussCtrl: Multi-View Consistent Text-Driven 3D Gaussian Splatting Editing}},
journal = {ECCV},
year = {2024},
}

@misc{edstedt2023romarobustdensefeature,
      title={RoMa: Robust Dense Feature Matching}, 
      author={Johan Edstedt and Qiyu Sun and Georg Bökman and Mårten Wadenbäck and Michael Felsberg},
      year={2023},
      eprint={2305.15404},
      archivePrefix={arXiv},
      primaryClass={cs.CV},
      url={https://arxiv.org/abs/2305.15404}, 
}

@article{bar2023multidiffusion,
  title={MultiDiffusion: Fusing Diffusion Paths for Controlled Image Generation},
  author={Bar-Tal, Omer and Yariv, Lior and Lipman, Yaron and Dekel, Tali},
  journal={arXiv preprint arXiv:2302.08113},
  year={2023}
}

@misc{alaluf2023crossimage,
    title={Cross-Image Attention for Zero-Shot Appearance Transfer}, 
    author={Yuval Alaluf and Daniel Garibi and Or Patashnik and Hadar Averbuch-Elor and Daniel Cohen-Or},
    year={2023},
    eprint={2311.03335},
    archivePrefix={arXiv},
    primaryClass={cs.CV}
}

@InProceedings{Cao_2023_ICCV,
  author    = {Cao, Mingdeng and Wang, Xintao and Qi, Zhongang and Shan, Ying and Qie, Xiaohu and Zheng, Yinqiang},
  title     = {MasaCtrl: Tuning-Free Mutual Self-Attention Control for Consistent Image Synthesis and Editing},
  booktitle = {Proceedings of the IEEE/CVF International Conference on Computer Vision (ICCV)},
  month     = {October},
  year      = {2023},
  pages     = {22560-22570}
}

@inproceedings{
  tewel2025addit,
  title={Add-it: Training-Free Object Insertion in Images With Pretrained Diffusion Models},
  author={Yoad Tewel and Rinon Gal and Dvir Samuel and Yuval Atzmon and Lior Wolf and Gal Chechik},
  booktitle={The Thirteenth International Conference on Learning Representations},
  year={2025},
  url={https://openreview.net/forum?id=ZeaTvXw080}
}

@article{shin2024diptychprompting,
  title={Large-Scale Text-to-Image Model with Inpainting is a Zero-Shot Subject-Driven Image Generator},
  author={Chaehun Shin and Jooyoung Choi and Heeseung Kim and Sungroh Yoon},
  booktitle={arXiv preprint arxiv:2411.15466},
  year={2024}
}

@article{epstein2023selfguidance,
  title={Diffusion Self-Guidance for Controllable Image Generation},
  author={Epstein, Dave and Jabri, Allan and Poole, Ben and Efros, Alexei A. and Holynski, Aleksander},
  booktitle={Advances in Neural Information Processing Systems},
  year={2023}
}

@article{shi2023dragdiffusion,
  title={DragDiffusion: Harnessing Diffusion Models for Interactive Point-based Image Editing},
  author={Shi, Yujun and Xue, Chuhui and Pan, Jiachun and Zhang, Wenqing and Tan, Vincent YF and Bai, Song},
  journal={arXiv preprint arXiv:2306.14435},
  year={2023}
}

@misc{chefer2023attendandexciteattentionbasedsemanticguidance,
      title={Attend-and-Excite: Attention-Based Semantic Guidance for Text-to-Image Diffusion Models}, 
      author={Hila Chefer and Yuval Alaluf and Yael Vinker and Lior Wolf and Daniel Cohen-Or},
      year={2023},
      eprint={2301.13826},
      archivePrefix={arXiv},
      primaryClass={cs.CV},
      url={https://arxiv.org/abs/2301.13826}, 
}

@article{kumari2022customdiffusion,
  title={Multi-Concept Customization of Text-to-Image Diffusion},
  author={Kumari, Nupur and Zhang, Bingliang and Zhang, Richard and Shechtman, Eli and Zhu, Jun-Yan},
  booktitle = {Proceedings of the IEEE/CVF Conference on Computer Vision and Pattern Recognition (CVPR)},
  year      = {2023}
}

@inproceedings{ruiz2023dreambooth,
  title={Dreambooth: Fine tuning text-to-image diffusion models for subject-driven generation},
  author={Ruiz, Nataniel and Li, Yuanzhen and Jampani, Varun and Pritch, Yael and Rubinstein, Michael and Aberman, Kfir},
  booktitle={Proceedings of the IEEE/CVF Conference on Computer Vision and Pattern Recognition},
  year={2023}
}

@article{barron2021mipnerf,
    title={Mip-NeRF: A Multiscale Representation 
           for Anti-Aliasing Neural Radiance Fields},
    author={Jonathan T. Barron and Ben Mildenhall and 
            Matthew Tancik and Peter Hedman and 
            Ricardo Martin-Brualla and Pratul P. Srinivasan},
    journal={ICCV},
    year={2021}
}

@misc{li2022neural3dvideosynthesis,
      title={Neural 3D Video Synthesis from Multi-view Video}, 
      author={Tianye Li and Mira Slavcheva and Michael Zollhoefer and Simon Green and Christoph Lassner and Changil Kim and Tanner Schmidt and Steven Lovegrove and Michael Goesele and Richard Newcombe and Zhaoyang Lv},
      year={2022},
      eprint={2103.02597},
      archivePrefix={arXiv},
      primaryClass={cs.CV},
      url={https://arxiv.org/abs/2103.02597}, 
}

@misc{labs2025flux1kontextflowmatching,
      title={FLUX.1 Kontext: Flow Matching for In-Context Image Generation and Editing in Latent Space}, 
      author={Black Forest Labs and Stephen Batifol and Andreas Blattmann and Frederic Boesel and Saksham Consul and Cyril Diagne and Tim Dockhorn and Jack English and Zion English and Patrick Esser and Sumith Kulal and Kyle Lacey and Yam Levi and Cheng Li and Dominik Lorenz and Jonas Müller and Dustin Podell and Robin Rombach and Harry Saini and Axel Sauer and Luke Smith},
      year={2025},
      eprint={2506.15742},
      archivePrefix={arXiv},
      primaryClass={cs.GR},
      url={https://arxiv.org/abs/2506.15742}, 
}

@misc{simeoni2025dinov3,
      title={DINOv3}, 
      author={Oriane Siméoni and Huy V. Vo and Maximilian Seitzer and Federico Baldassarre and Maxime Oquab and Cijo Jose and Vasil Khalidov and Marc Szafraniec and Seungeun Yi and Michaël Ramamonjisoa and Francisco Massa and Daniel Haziza and Luca Wehrstedt and Jianyuan Wang and Timothée Darcet and Théo Moutakanni and Leonel Sentana and Claire Roberts and Andrea Vedaldi and Jamie Tolan and John Brandt and Camille Couprie and Julien Mairal and Hervé Jégou and Patrick Labatut and Piotr Bojanowski},
      year={2025},
      eprint={2508.10104},
      archivePrefix={arXiv},
      primaryClass={cs.CV},
      url={https://arxiv.org/abs/2508.10104}, 
}

@inproceedings{fu2023dreamsim,
    title={DreamSim: Learning New Dimensions of Human Visual Similarity using Synthetic Data},
    author= {Fu, Stephanie and Tamir, Netanel and Sundaram, Shobhita and Chai, Lucy and Zhang, Richard and Dekel, Tali and Isola, Phillip},
    booktitle={Advances in Neural Information Processing Systems},
    pages={50742--50768},
    volume={36},
    year={2023}
}

@inproceedings{avrahami2024chosen,
  author = {Avrahami, Omri and Hertz, Amir and Vinker, Yael and Arar, Moab and Fruchter, Shlomi and Fried, Ohad and Cohen-Or, Daniel and Lischinski, Dani},
  title = {The Chosen One: Consistent Characters in Text-to-Image Diffusion Models},
  year = {2024},
  isbn = {9798400705250},
  publisher = {Association for Computing Machinery},
  address = {New York, NY, USA},
  url = {https://doi.org/10.1145/3641519.3657430},
  doi = {10.1145/3641519.3657430},
  booktitle = {ACM SIGGRAPH 2024 Conference Papers},
  articleno = {26},
  numpages = {12},
  location = {Denver, CO, USA},
  series = {SIGGRAPH '24}
}

@misc{zer0int_LongCLIP_L_Diffusers,
  author       = {zer0int},
  title        = {LongCLIP‐L‐Diffusers},
  howpublished = {\url{https://huggingface.co/zer0int/LongCLIP‐L‐Diffusers}},
  note         = {Model on Hugging Face. Accessed: 2025-10-18},
  year         = {2024}
}

@techreport{openai2025gpt5,
  title        = {GPT-5 System Card},
  author       = {OpenAI},
  year         = {2025},
  month        = {August},
  institution  = {OpenAI},
  url          = {https://cdn.openai.com/gpt-5-system-card.pdf},
  note         = {Technical report}
}

@techreport{openai2024gpt4o,
  title        = {GPT-4o System Card},
  author       = {OpenAI},
  year         = {2024},
  month        = {May},
  institution  = {OpenAI},
  url          = {https://cdn.openai.com/gpt-4o-system-card.pdf},
  note         = {Technical report}
}

@article{zhou2024storydiffusion,
  title={StoryDiffusion: Consistent Self-Attention for Long-Range Image and Video Generation},
  author={Zhou, Yupeng and Zhou, Daquan and Cheng, Ming-Ming and Feng, Jiashi and Hou, Qibin},
  journal={NeurIPS 2024},
  year={2024}
}

@InProceedings{Liu_2024_CVPR,
    author    = {Liu, Chang and Wu, Haoning and Zhong, Yujie and Zhang, Xiaoyun and Wang, Yanfeng and Xie, Weidi},
    title     = {Intelligent Grimm - Open-ended Visual Storytelling via Latent Diffusion Models},
    booktitle = {Proceedings of the IEEE/CVF Conference on Computer Vision and Pattern Recognition (CVPR)},
    month     = {June},
    year      = {2024},
    pages     = {6190-6200}
}

@misc{he2024dreamstory,
  title={DreamStory: Open-Domain Story Visualization by LLM-Guided Multi-Subject Consistent Diffusion}, 
  author={Huiguo He and Huan Yang and Zixi Tuo and Yuan Zhou and Qiuyue Wang and Yuhang Zhang and Zeyu Liu and Wenhao Huang and Hongyang Chao and Jian Yin},
  year={2024},
  eprint={2407.12899},
  archivePrefix={arXiv},
  primaryClass={cs.CV},
  url={https://arxiv.org/abs/2407.12899}, 
}

@article{ye2023ip-adapter,
  title={IP-Adapter: Text Compatible Image Prompt Adapter for Text-to-Image Diffusion Models},
  author={Ye, Hu and Zhang, Jun and Liu, Sibo and Han, Xiao and Yang, Wei},
  booktitle={arXiv preprint arxiv:2308.06721},
  year={2023}
}

@software{google2025nanobanana,
  author       = {Google DeepMind},
  title        = {Gemini 2.5 Flash Image (“Nano Banana”) model/API},
  year         = {2025},
  note         = {Accessible via Google Gemini API},
}

@article{mou2023t2i,
  title={T2i-adapter: Learning adapters to dig out more controllable ability for text-to-image diffusion models},
  author={Mou, Chong and Wang, Xintao and Xie, Liangbin and Wu, Yanze and Zhang, Jian and Qi, Zhongang and Shan, Ying and Qie, Xiaohu},
  journal={arXiv preprint arXiv:2302.08453},
  year={2023}
}

@InProceedings{brooks2022instructpix2pix,
    author     = {Brooks, Tim and Holynski, Aleksander and Efros, Alexei A.},
    title      = {InstructPix2Pix: Learning to Follow Image Editing Instructions},
    booktitle  = {CVPR},
    year       = {2023},
}

@article{song2020denoising,
  title={Denoising Diffusion Implicit Models},
  author={Song, Jiaming and Meng, Chenlin and Ermon, Stefano},
  journal={arXiv:2010.02502},
  year={2020},
  month={October},
  abbr={Preprint},
  url={https://arxiv.org/abs/2010.02502}
}

@article{hu2022lora,
  title={Lora: Low-rank adaptation of large language models.},
  author={Hu, Edward J and Shen, Yelong and Wallis, Phillip and Allen-Zhu, Zeyuan and Li, Yuanzhi and Wang, Shean and Wang, Lu and Chen, Weizhu and others},
  journal={ICLR},
  volume={1},
  number={2},
  pages={3},
  year={2022}
}

@InProceedings{Tumanyan_2023_CVPR,
    author    = {Tumanyan, Narek and Geyer, Michal and Bagon, Shai and Dekel, Tali},
    title     = {Plug-and-Play Diffusion Features for Text-Driven Image-to-Image Translation},
    booktitle = {Proceedings of the IEEE/CVF Conference on Computer Vision and Pattern Recognition (CVPR)},
    month     = {June},
    year      = {2023},
    pages     = {1921-1930}
}

@article{hertz2022prompt,
  title={Prompt-to-prompt image editing with cross attention control},
  author={Hertz, Amir and Mokady, Ron and Tenenbaum, Jay and Aberman, Kfir and Pritch, Yael and Cohen-Or, Daniel},
  booktitle={arXiv preprint arXiv:2208.01626},
  year={2022}
}

@article{zhu2025kv,
  title={KV-Edit: Training-Free Image Editing for Precise Background Preservation},
  author={Zhu, Tianrui and Zhang, Shiyi and Shao, Jiawei and Tang, Yansong},
  journal={arXiv preprint arXiv:2502.17363},
  year={2025}
}

@inproceedings{yuan2024instructvideo,
  title={Instructvideo: Instructing video diffusion models with human feedback},
  author={Yuan, Hangjie and Zhang, Shiwei and Wang, Xiang and Wei, Yujie and Feng, Tao and Pan, Yining and Zhang, Yingya and Liu, Ziwei and Albanie, Samuel and Ni, Dong},
  booktitle={Proceedings of the IEEE/CVF Conference on Computer Vision and Pattern Recognition},
  pages={6463--6474},
  year={2024}
}

@misc{prolific2024,
  author       = {Prolific},
  title        = {Prolific},
  year         = {2024},
  howpublished = {\url{https://www.prolific.com/}}
}

@misc{wan2025wanopenadvancedlargescale,
      title={Wan: Open and Advanced Large-Scale Video Generative Models}, 
      author={Team Wan and Ang Wang and Baole Ai and Bin Wen and Chaojie Mao and Chen-Wei Xie and Di Chen and Feiwu Yu and Haiming Zhao and Jianxiao Yang and Jianyuan Zeng and Jiayu Wang and Jingfeng Zhang and Jingren Zhou and Jinkai Wang and Jixuan Chen and Kai Zhu and Kang Zhao and Keyu Yan and Lianghua Huang and Mengyang Feng and Ningyi Zhang and Pandeng Li and Pingyu Wu and Ruihang Chu and Ruili Feng and Shiwei Zhang and Siyang Sun and Tao Fang and Tianxing Wang and Tianyi Gui and Tingyu Weng and Tong Shen and Wei Lin and Wei Wang and Wei Wang and Wenmeng Zhou and Wente Wang and Wenting Shen and Wenyuan Yu and Xianzhong Shi and Xiaoming Huang and Xin Xu and Yan Kou and Yangyu Lv and Yifei Li and Yijing Liu and Yiming Wang and Yingya Zhang and Yitong Huang and Yong Li and You Wu and Yu Liu and Yulin Pan and Yun Zheng and Yuntao Hong and Yupeng Shi and Yutong Feng and Zeyinzi Jiang and Zhen Han and Zhi-Fan Wu and Ziyu Liu},
      year={2025},
      eprint={2503.20314},
      archivePrefix={arXiv},
      primaryClass={cs.CV},
      url={https://arxiv.org/abs/2503.20314}, 
}
}

\newpage
\appendix

\renewcommand{\thefigure}{\thesection.\arabic{figure}}
\setcounter{figure}{0}
\renewcommand{\theequation}{\thesection.\arabic{equation}}
\setcounter{equation}{0}
\renewcommand{\thetable}{\thesection.\arabic{table}}
\setcounter{table}{0}

\section*{Appendix}

\section{Implementation details}
\label{sec:impl}

\subsection{Method}
\label{sec:impl_method}

\begin{figure}[t!]
  \centering
  \includegraphics[width=1.0\linewidth,keepaspectratio]{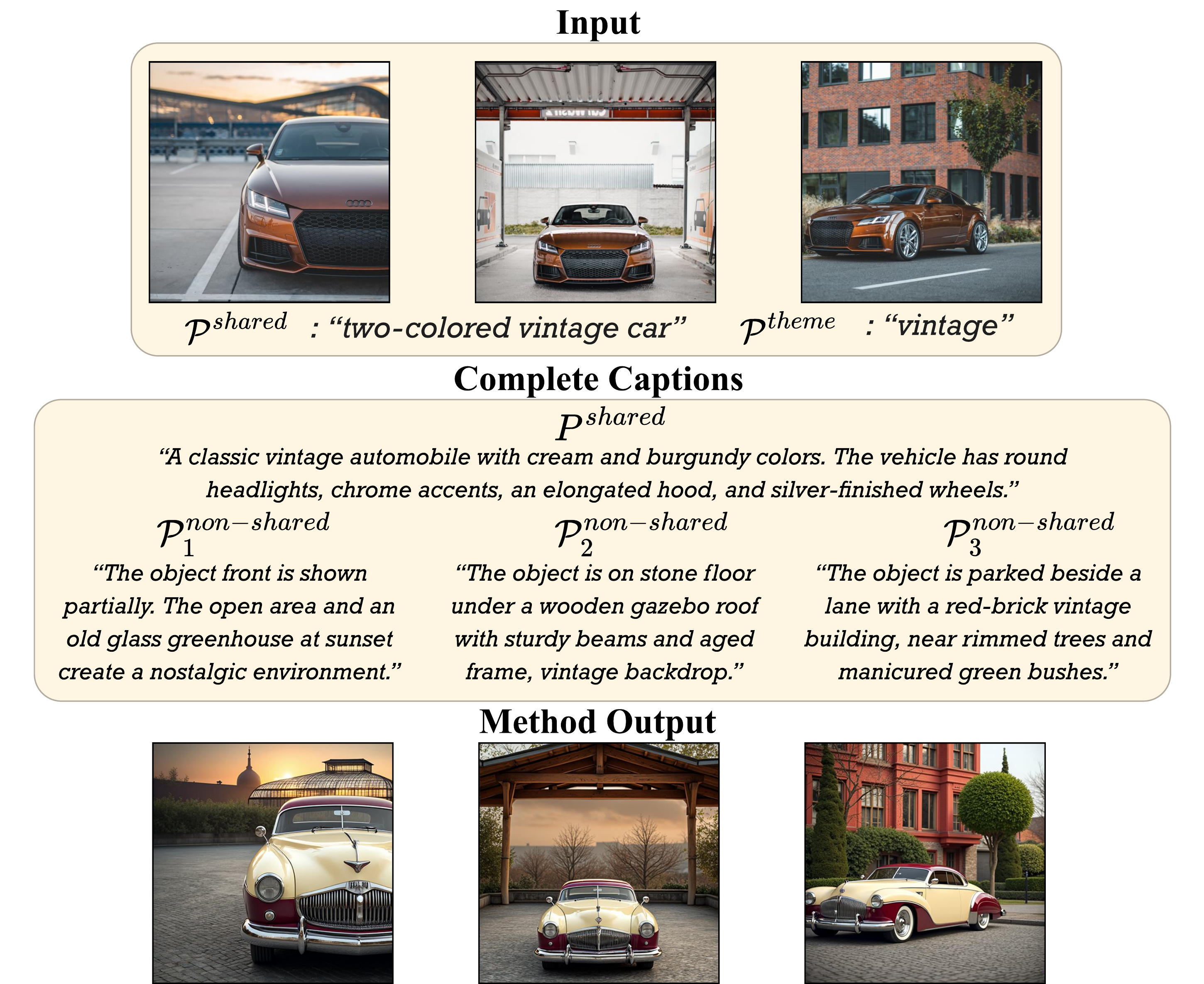}%
  \vspace{-7pt}
  \caption{\textbf{Completed captions.} 
  The VLM provides detailed per-image captions aligned with $\mathcal{P}^{\text{theme}}$ and expands on shared content \(\mathcal{P}^{\text{shared}}\), offering an alternative to manual prompting.
  }
  \label{fig:caption}
\end{figure}

\begin{figure*}[h]
  \centering
    \includegraphics[
    width=1.0 \linewidth,
    keepaspectratio
  ]{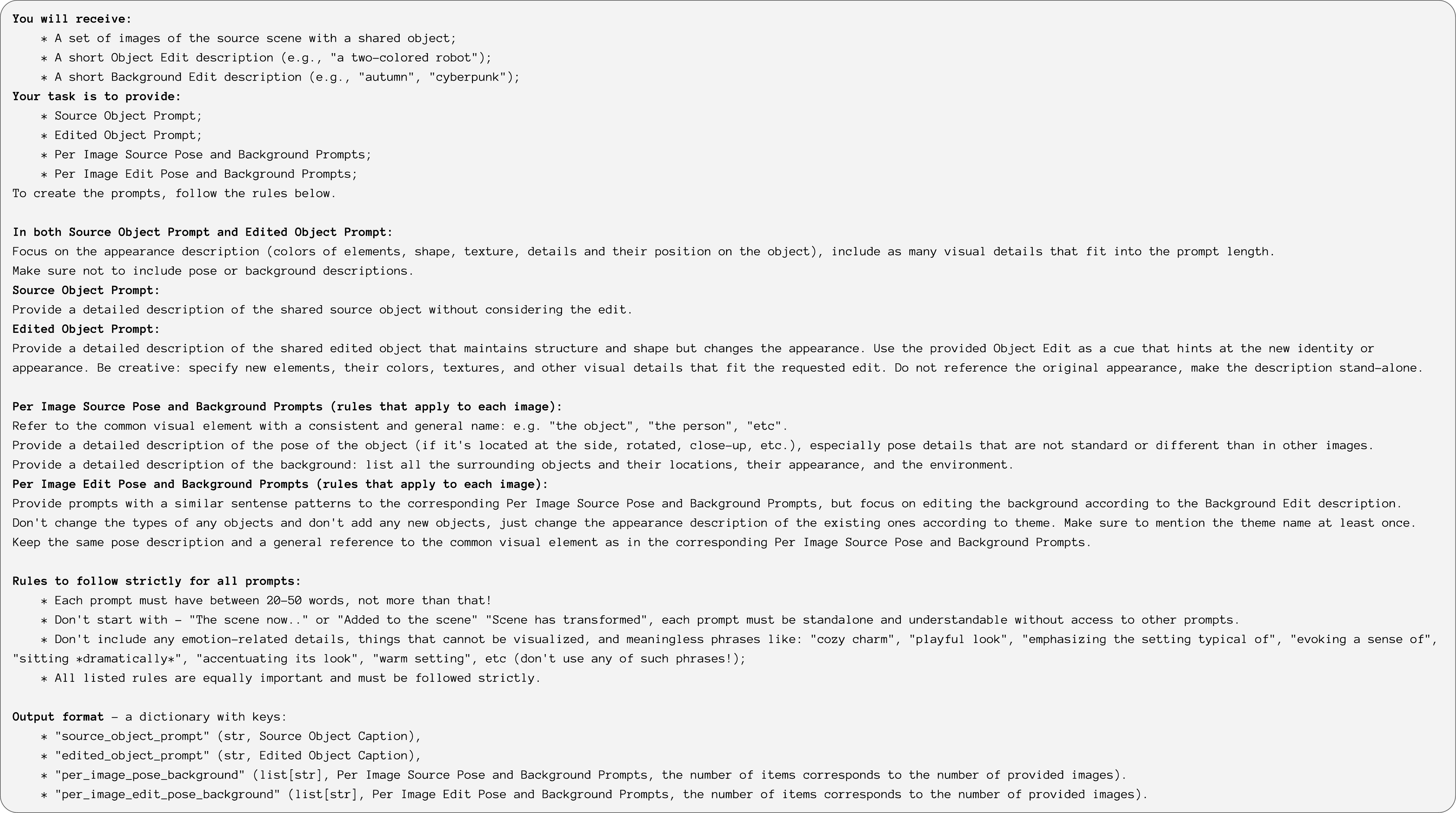}
  \vspace{-10pt}
  \caption{\textbf{VLM instructions for prompt construction.} System prompt. Source prompts are used for localized editing with FlowEdit.}
  \label{fig:sys_prompts} \afterfigure
\end{figure*}

\textit{Preprocessing.}
Prompt construction with a VLM (\cref{sec:prompting}) is used as an automated, scalable alternative to manual prompting. We instruct GPT-4o \cite{openai2024gpt4o} via the system prompt (\cref{fig:sys_prompts}) to generate $P^{non-shared}_i$ and enrich $\mathcal{P}^\text{shared}$ with creative artistic details (e.g., colors, textures). \Cref{fig:caption} exemplifies VLM input and output. Source 2D correspondences and shared content are identified using RoMA \cite{edstedt2023romarobustdensefeature} predictions filtered by a confidence threshold $\operatorname{c} > 0.05$.

\textit{General.} All results are generated using FLUX \cite{flux2024} with $t=25\dots0$ steps. We apply ControlNet \cite{fluxcontrolnet2024} complementing DepthAnythingV2 \cite{yang2024depthv2} depth with HED edge maps \cite{xie2015holisticallynestededgedetection} with equal weights, and linearly anneal control strength $w_{\text{ctrl}}^t=\text{linear\_anneal}(1,0,t)$. This enforces correct layout while preventing over-constraining control.

\begin{figure}[t!]
  \begin{center}
    \includegraphics[%
    width=1.0\linewidth,%
  ]{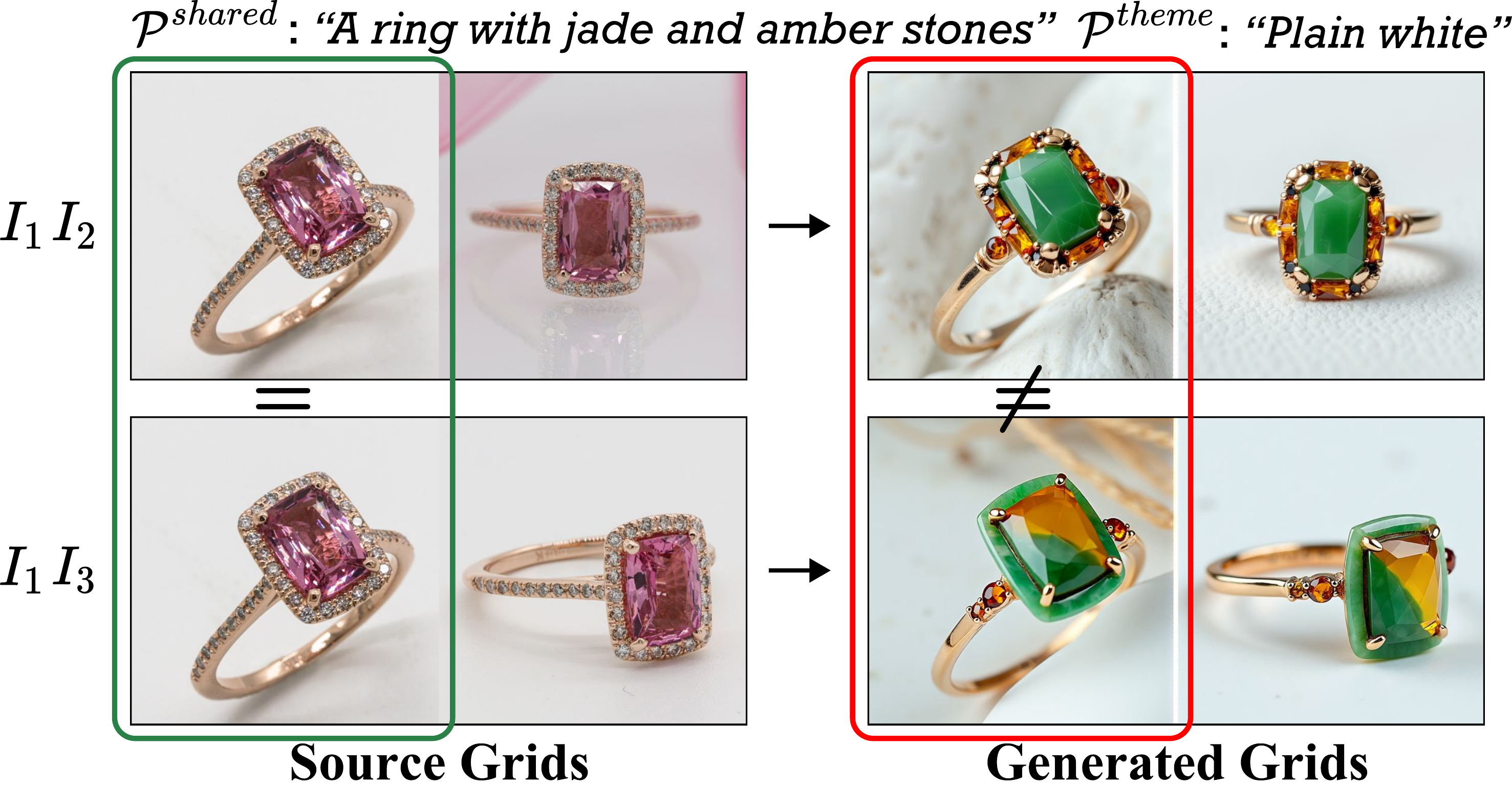}%
  \end{center}
  \vspace{-15pt}
  \caption{\textbf{Image versions.} 
  A grid of $I_1$ with varying neighbors produces pairs that are mutually consistent but exhibit different appearances across pairs. This appearance divergence is prevented every denoising step in \graph by a graph-wide MFF and version averaging.
  }\afterfigure
  \label{fig:versions}
\end{figure}

In \textit{\graph} (\cref{sec:graph}), independent Gaussian noise is sampled for each node. Two-image grids are concatenated horizontally or vertically, depending on the image aspect ratio.
For $\mathcal{P}_{ij}$, any freely written prefix prompting a side-by-side layout yields similar results. \cref{fig:versions} shows the concept of image versions aggregated along graph edges and unified with MFF.

\textit{\mff} (\cref{sec:mff}) and \textit{\mguide} (\cref{sec:guide}) are applied to keys and values before RoPE.
MFF operates on all double-stream blocks for $t^{MFF}_\text{d} > 3$ and the last three single-stream blocks for $t^{MFF}_\text{s} > 15$.
\mguide objective is defined on all blocks for $t^{guide}_\text{a} > 10$. Latents are updated using Adam with a linearly annealed learning rate $\alpha^t = \operatorname{linear\_anneal}(0.016, 0.002, t)$ and gradient accumulation per edge.
In \cref{eq:mff_graph}, undefined $M_{ij}(\mathbf{c})$ terms are omitted from averaging.

\textit{Localized editing.} 
FlowEdit \cite{kulikov2024flowedit} is an inversion-free method that enables background-preserving editing by traversing a path between source and target image distributions with $v$-predictions $V = V^{tar}(z^{tar}) - V^{src}(z^{src})$. To incorporate it into our method, we redefine $V_{ij}$ implied in $\operatorname{denoise}_{\operatorname{MFF}}(\cdot)$ in \cref{eq:graph_denoise} as $V_{ij} = V^{tar}_{ij} - V^{src}_{ij}$, where
{
\setlength{\abovedisplayskip}{6pt}
\setlength{\belowdisplayskip}{6pt}
\begin{equation}
\scalebox{1.}{$
\displaystyle
V^{\alpha}_{ij} := V^{\alpha}_{ij}(\{z^{\alpha}_{ij}\}_{e \in E}, \{\mathcal{P}^{\alpha}_{ij}\}_{e \in E}, \{M_{ij}\}_{e \in E}),
$}
\end{equation}}\vspace{-13pt}
{
\setlength{\abovedisplayskip}{6pt}
\setlength{\belowdisplayskip}{6pt}
\begin{equation}
\scalebox{1.}{$
\displaystyle
\alpha \in \{src, tar\}
$}\vspace{-7pt}
\end{equation}}
Similar to \cite{meng2022sdeditguidedimagesynthesis}, FlowEdit relies on a single starting timestep parameter $n_{\text{max}}$ to balance structure preservation and appearance deviation. 
We ease this trade-off by adopting spatial control with an earlier start step.
The used hyperparameters are: $T=25, n_{\text{max}}=24, n_{\text{min}}=10, w_{\text{ctrl}}=0.5, t^{MFF}_\text{d} > 20$. \mguide is omitted. 

Despite the improved balance, some trade-off between structure and appearance remains. We therefore suggest $w_{\text{ctrl}}=0.8$ when structure is prioritized over editability.

\subsection{Baselines}
\label{sec:impl_base}

For \textit{IC-LoRA} \cite{lhhuang2024iclora}, we use the most general Film Storyboard checkpoint with a LoRA weight of 0.8. \textit{IC-LoRA} and \textit{FLUX} \cite{flux2024} are run with identical control signals as the method. \textit{FLUX} and \textit{FLUX Kontext} \cite{labs2025flux1kontextflowmatching} share the same random seed across images, and \textit{FLUX Kontext} is prompted by the instruction \textit{"Turn the $<$object$>$ into ..."}. For \textit{Edicho}~\cite{bai2024edicho}, we use the first image as a reference for the rest. Metrics in \cref{tab:num_eval,tab:win_rate} are computed with its public inpainting implementation available at publication, using the shared region masks from evaluation. The results in \cref{fig:qualitative} with the ControlNet pipeline were provided by the authors.

\subsection{Evaluation Setup}
\label{sec:impl_metrics}

\textit{CLIP score} is computed with LongCLIP-L \cite{zer0int_LongCLIP_L_Diffusers} and prompts \textit{"[$\mathcal{P}^{shared}$]. [$\mathcal{P}_i^{non-shared}$]."}. 

\begin{figure*}[ht]
  \centering
  \includegraphics[width=\linewidth,keepaspectratio]{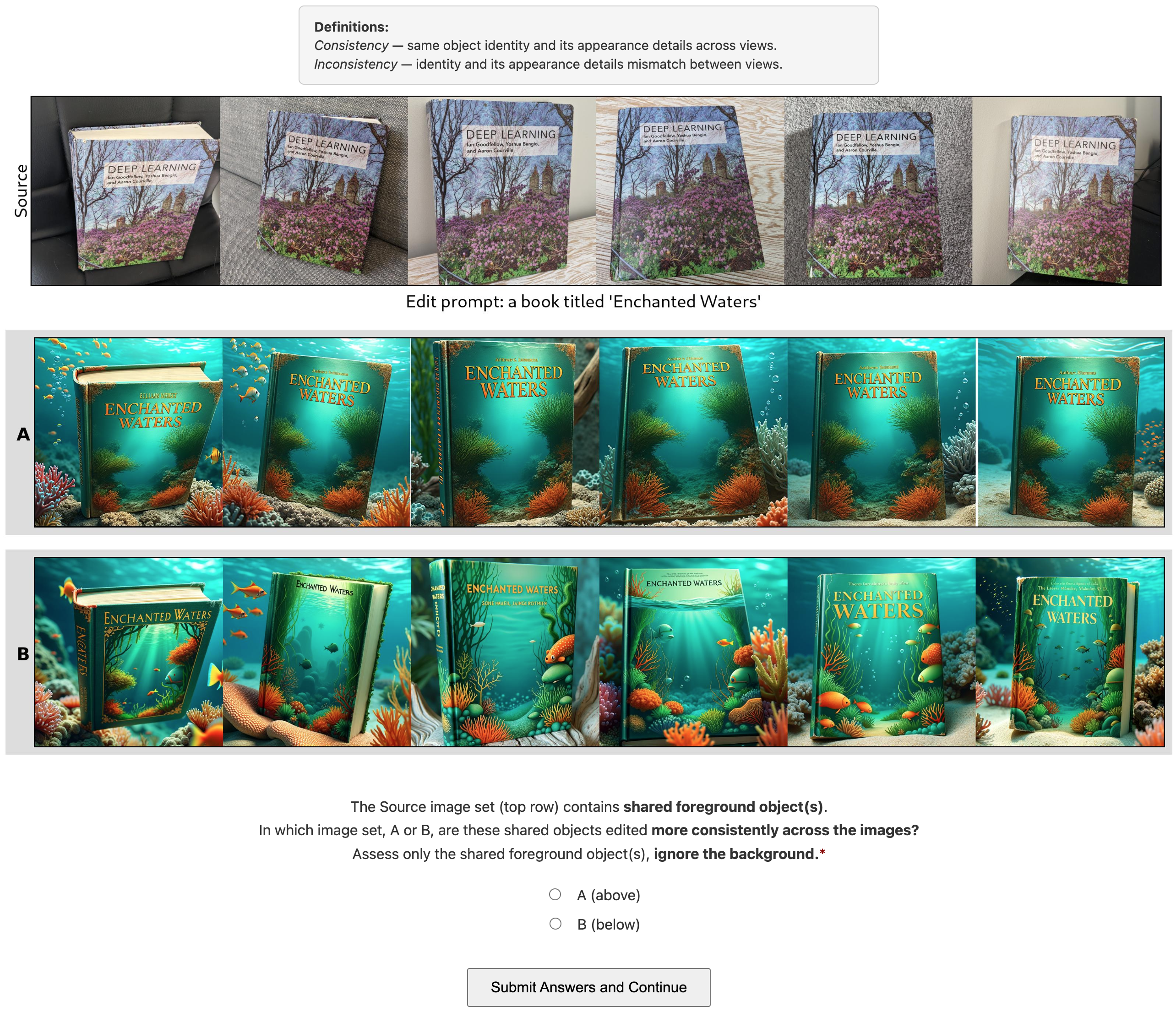}
  \vspace{-15pt}
  \caption{\textbf{User study interface.} Example of an interface with a question comparing \method to FLUX \cite{flux2024}.} \afterfigure
  \label{fig:study_example}
\end{figure*}

The \textit{User Study} evaluating consistency was conducted on 80 random edits from our benchmark using the Prolific \cite{prolific2024} platform. Participants were shown the source image set, $\mathcal{P}^{shared}$, outputs from \method and a baseline in a randomized 2AFC setup (see the interface in \cref{fig:study_example}). They answered: \textit{"The source image set contains shared foreground object(s). In which image set, A or B, are these objects edited more consistently across images?"}. We collected 1,600 judgments from 40 participants, with 5 respondents per comparison.

The automatic \textit{VLM evaluation} was conducted on the full benchmark using GPT-5 \cite{openai2025gpt5} with the system prompt: \textit{"You will receive an image to evaluate along with an evaluation question. Output A or B."}. Each comparison used the same inputs as human participants: the compiled image (source set with randomized generations A and B), $\mathcal{P}^{shared}$, and the user-study question.

Our \textit{DINO-MatchSim} metric quantifies fine-grained cross-image consistency in appearance and structure. Generated images are compared in the DINOv3 \cite{simeoni2025dinov3} feature space at fixed correspondences $\textit{NN}_{ij}$—nearest neighbors from background-removed source images. DINOv3 is used for its strong and versatile representation and to ensure independence from the 2D matches used by our method.
In practice, pairwise cosine similarities $S_{ij}$ (see \cref{eq:dino_matchsim}) are averaged into $\bar{S}$, then transformed as: $\textit{DINO-MatchSim} = \exp((\bar{S} - 1)/\tau)$ with $\tau=0.6$. This mapping amplifies consistency differences by compressing negative similarities into small scores ($\approx 0–0.2$) and expanding positive similarities into a wider interval ($\approx 0.2-1$).

\Cref{tab:benchmark} shows the distribution of the evaluation set. 

\vspace{-1.5ex}
\setlength{\tabcolsep}{2pt}
\begin{table}[t]
  \centering
  \scalebox{0.85}{
  \begin{tabular}{lccp{3.6cm}}
    \toprule
    Shared Content & \# Sets & \# Edits & \multicolumn{1}{c}{Source Datasets} \\
    \midrule
    Subject-driven & 131 & 357 &
      \shortstack[l]{CustomConcepts101 \cite{kumari2022customdiffusion}\\
                     DreamBooth \cite{ruiz2023dreambooth}} \\
    \midrule
    Subject \& Background & 11 & 27 &
      \shortstack[l]{DyNeRF \cite{li2022neural3dvideosynthesis}\\
                     Instruct-NeRF2NeRF \cite{instructnerf2023}\\
                     Mip-NeRF \cite{barron2021mipnerf}\\
                     royalty-free videos} \\
    \midrule
    Storyboard sketch & 7 & 16 & ChatGPT-generated \\
    \bottomrule
  \end{tabular}
  }
  \vspace{-0.2cm}
  \caption{\textbf{Benchmark composition.} A distribution of sets and edits by the types of shared content.}
  \label{tab:benchmark}
  \vspace{-12pt}
\end{table}

\section{Runtime \& Memory}
\label{sec:runtime}

\begin{figure}[t!]
  \begin{center}
    \includegraphics[%
    width=1.0\linewidth,%
  ]{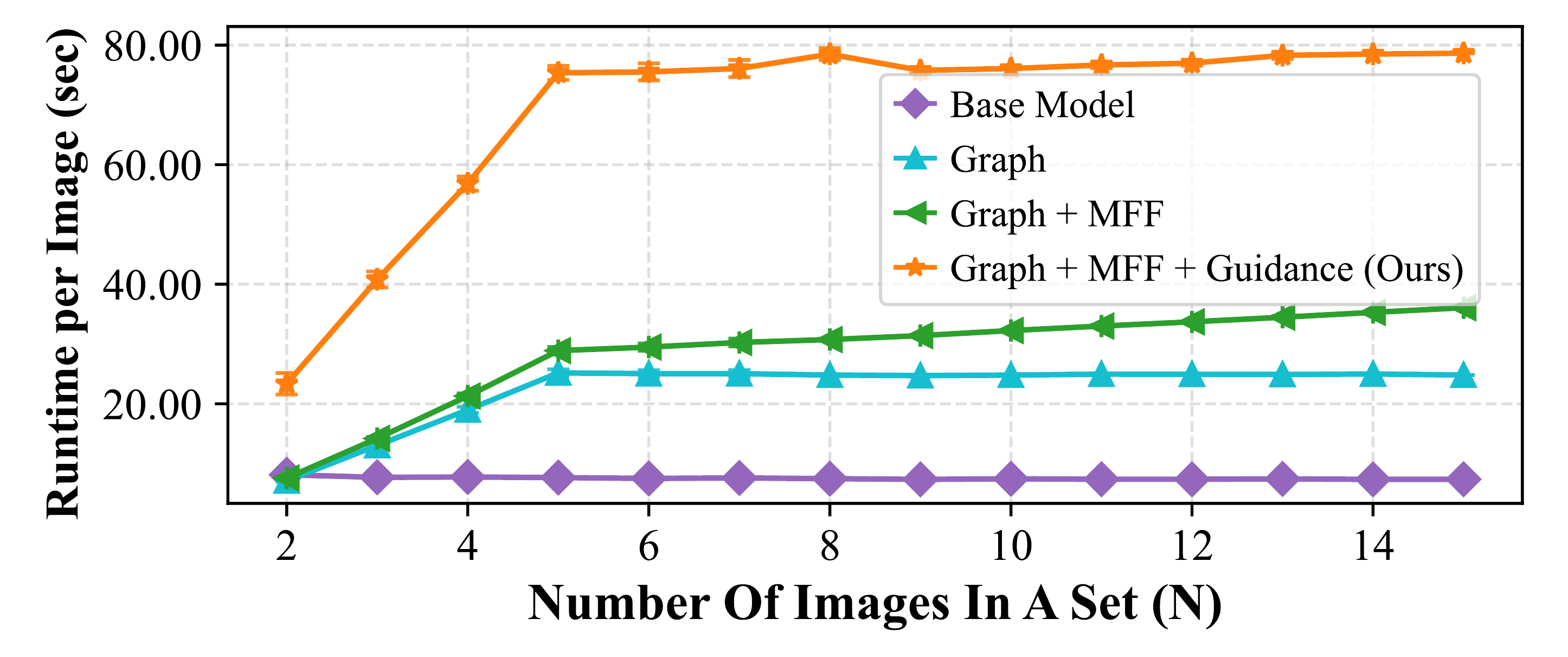}%
  \end{center}
  \vspace{-15pt}
  \caption{\textbf{Runtime per Image vs. Set Size.} 
  \method maintains approximately constant runtime per image for $N \geq 5$.
  } \afterfigure
  \label{fig:runtime}
\end{figure}

During the extraction of 2D matches in preprocessing, RoMA has a runtime of 198.8~ms per image pair on an RTX6000 GPU \cite{edstedt2023romarobustdensefeature}.
In \cref{fig:runtime}, we report the per-image runtime of our full method as a function of $N$, progressively adding its components. As also discussed in \cref{sec:analysis}, \method maintains linear runtime for $N \geq 5$ by constraining graph connectivity to a node degree of 4. Since \mguide noticeably increases runtime, it can be omitted when efficiency is critical, trading fine-grained consistency for speed. 

\begin{figure*}[ht]
  \centering
  \includegraphics[width=\linewidth,keepaspectratio]{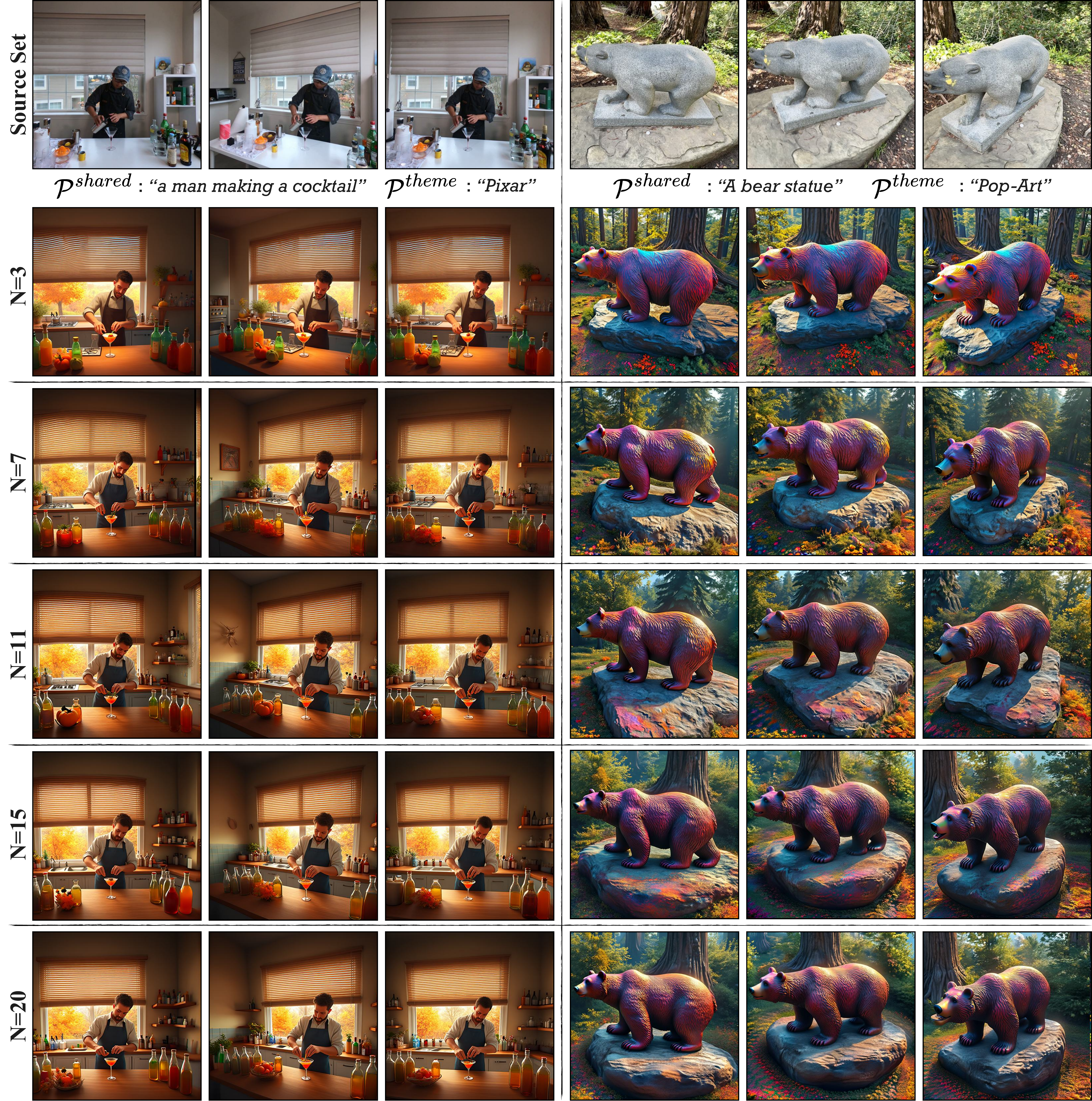}
  \vspace{-15pt}
  \caption{\textbf{Consistency vs. Set Size.} The same 3 images when generated as part of increasingly bigger sets of size $N$. Low and middle frequencies remain consistent, while high frequencies disagree more with higher $N$.} \afterfigure
  \label{fig:num_img_mem}
\end{figure*}

In terms of memory constraints, we can generate up to 20 images using the full method, or 22 without \mguide. 
\Cref{fig:num_img_mem} illustrates how increasing the set size to the maximum affects consistency.

All measurements are conducted on an NVIDIA A100 GPU with an image resolution of $512 \times 512$. 

\section{Adapting Extended Attention for Our Task}
\label{sec:ext_attn_appendix}

\begin{figure*}[ht]
  \centering
    \includegraphics[%
    width=1.0 \linewidth,%
    keepaspectratio%
  ]{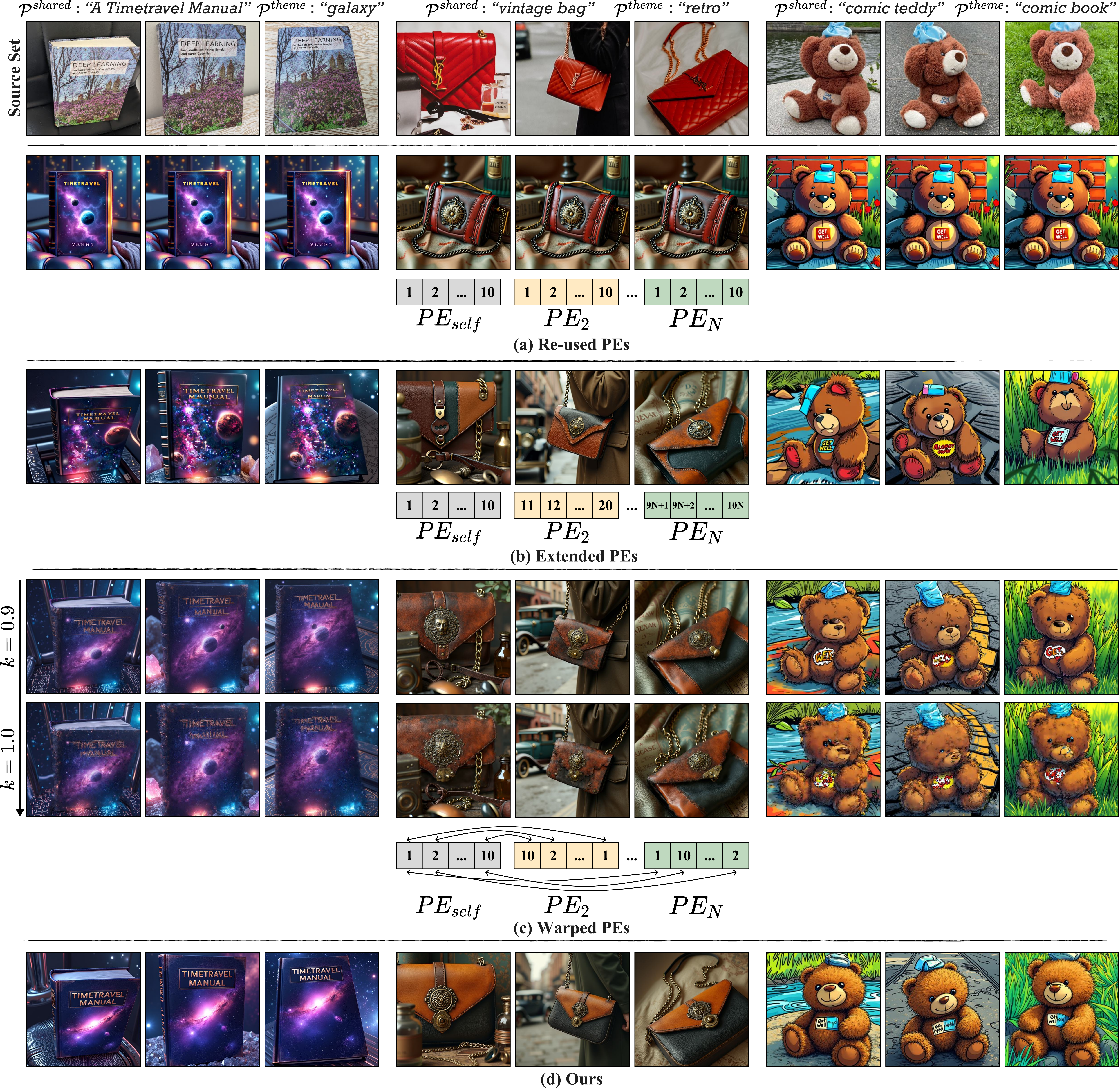}%
  \vspace{-10pt}
  \caption{\textbf{Extended Attention in DiTs.} (a-c) Settings varying by PEs of extra keys, illustrated in 1D for simplicity below each setting, (d) Ours. (a) Re-using PEs leads to duplicated generations. (b) Extending PEs leads to low consistency and artifacts. (c) Warping PEs by source matches demonstrates a severe quality-consistency tradeoff.}
  \label{fig:ext_attn} \afterfigure
\end{figure*}

We explore the additional qualitative baseline of prompting consistency through extended attention, widely used in UNets for cross-sample appearance sharing \cite{alaluf2023crossimage,Cao_2023_ICCV,tokenflow2023,tewel2024training}, and observe that in FLUX, this approach leads to artifacts and lacks semantic reasoning. We share our findings to inform future research. 

In DiTs, Rotary Position Embeddings (RoPE) are applied before self-attention in each block, thus extended keys of the image modality are:
{
\setlength{\abovedisplayskip}{2pt}
\setlength{\belowdisplayskip}{2pt}
\begin{equation}
\scalebox{0.90}{$
\displaystyle
K_{img} = [PE_{self}(K_{I^{self}}), PE_2(K_{I^2}), \dots, PE_N(K_{I^N})]
$}
\end{equation}}
The positional embeddings (PEs) with a 2D token coordinate must be defined for all keys. \cite{tewel2025addit} showed that in FLUX they strongly influence the spatial object placement. However, since extra keys are not truly present in the generated image plane, the choice of their positions is ambiguous. We test three different strategies: (i) Re-using $PE_{self}$. This variant yields identical generations that completely disregard the control signal (\cref{fig:ext_attn}a);
(ii) Continuing positions analogous to PEs of spatially concatenated images in the grid generation. Despite this similarity, without a prompt indicating a joint canvas, this setting leads to artifacts and fails to achieve consistent outputs (\cref{fig:ext_attn}b);
(iii) We extend only with keys that have a source 2D match and warp PEs of $K_{I^{self}}$ such that matched keys share the same position. Additionally, extra keys are scaled by $k$ to control their temperature in softmax (\cref{fig:ext_attn}c). With $k=1$, generations approach consistency at the cost of severe quality degradation. Reducing $k$ alleviates the artifacts, but compromises consistency before image quality is fully restored. In contrast, our method achieves consistency while avoiding artifacts (\cref{fig:ext_attn}d).
These findings demonstrate a strong connection between features that share the same position in RoPE. Exploring this behavior may open future research directions.

\section{Downstream Applications}
\label{sec:downstream}
\vspace{-5pt}

\begin{figure}[H]
  \centering
  \includegraphics[width=1.0\linewidth]{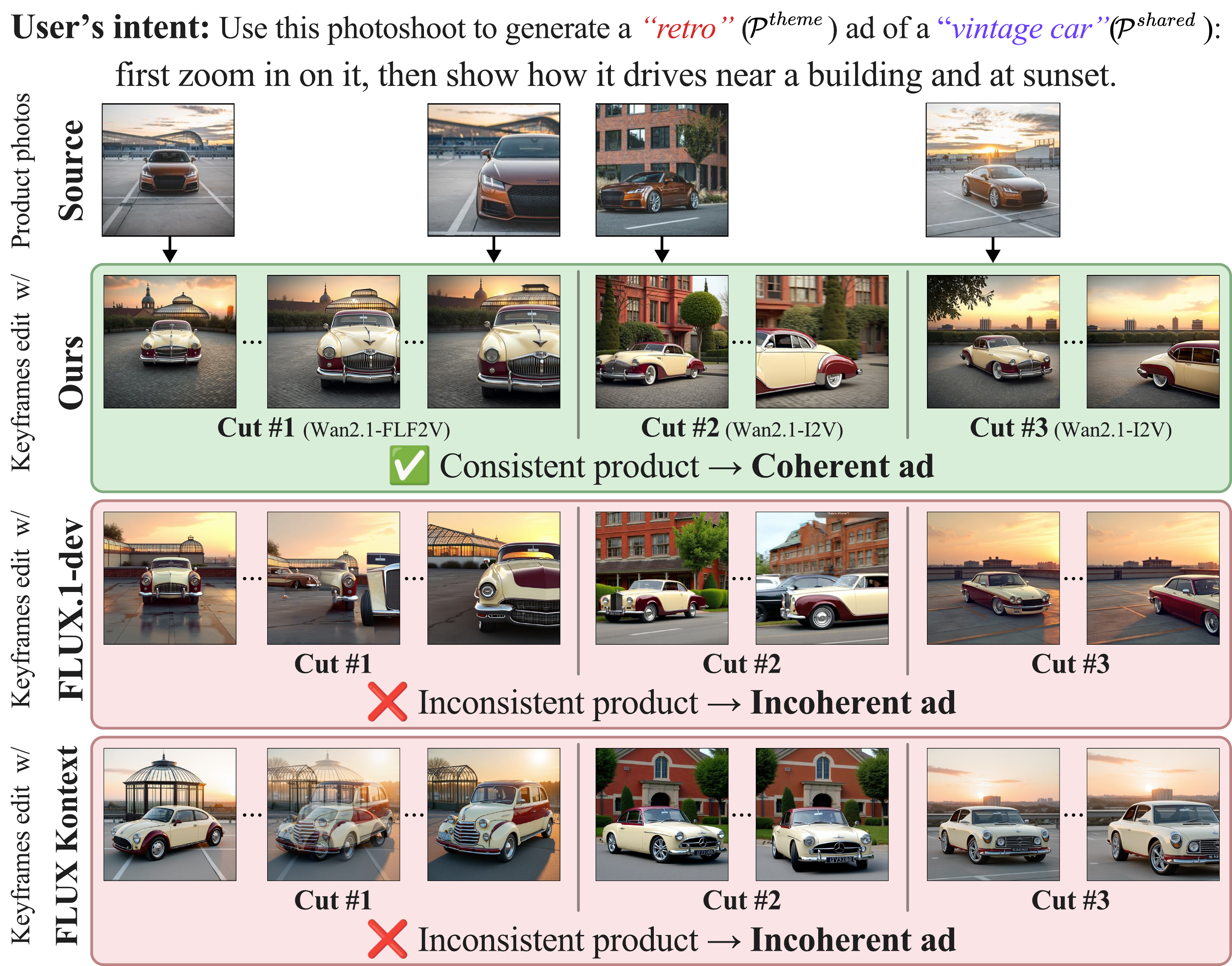}
   \caption{
   \textbf{Additional experiments: multi-cut ad editing.} 
   Product advertisements often depict multiple cuts of the same object in different contexts rather than one continuous shot, which can be obtained by conditioning an I2V model \cite{wan2025wanopenadvancedlargescale} 
   on keyframes. Our method enables infinite coherent edits with cross-cut consistency, where one-to-one methods fail.
   } \afterfigure
   \label{fig:video_edit}
\end{figure}

In addition to the extended applications in Sec. 3.3, in \cref{fig:video_edit} we demonstrate a downstream video editing application enabled by our method and not achievable with image-level editing alone, further motivating the task of consistent set-to-set editing.
\clearpage

\end{document}